\documentclass[a4paper,fleqn]{cas-sc}

\usepackage[authoryear,longnamesfirst]{natbib}
\usepackage{threeparttable}
\def\tsc#1{\csdef{#1}{\textsc{\lowercase{#1}}\xspace}}
\tsc{WGM}
\tsc{QE}
\usepackage{afterpage}
\usepackage{longtable}
\usepackage{caption}

\begin{document}
\let\WriteBookmarks\relax
\def\floatpagepagefraction{1}
\def\textpagefraction{.001}

\shorttitle{}    

\shortauthors{}  

\title [mode = title]{Incomplete Graph Learning: A Comprehensive Survey}


%


\author[1]{Riting Xia}[orcid=0000-0003-2184-4360
]

\ead{xiart19@mails.jlu.edu.cn}
\credit{Conceptualization, Investigation, Data curation, Methodology, Software, Validation, Visualization, Writing
}

\address[1]{College of Computer Science, Inner Mongolia University, Hohhot, 010021, China}

\author[1]{Huibo Liu}[orcid=0009-0005-7178-1419
]
\credit{Investigation, Resources, Visualization, Data curation, Writing}
\ead{liuhuibo@mail.imu.edu.cn}



\author[2,3]{Anchen Li}[orcid=0000-0001-9828-6964
       ]

\ead{liac@jlu.edu.cn}

\credit{Investigation, Resources, Writing - review \&
editing}

\author[2,3]{Xueyan Liu}[orcid=0000-0003-1790-3751
      ]
\ead{xueyanliu@jlu.edu.cn}


\author[1]{Yan Zhang}[orcid=0000-0002-5522-5851
]
\cormark[1]
\credit{Investigation, Resources, Validation, Writing - review \&
	editing}
\ead{yanz19@mails.jlu.edu.cn}

\author[2,3]{Chunxu Zhang}[orcid=0000-0003-0825-872X
      ]
\cormark[1]
\credit{Investigation, Methodology, Resources, Writing - review \&
editing}
\ead{zhangchunxu@jlu.edu.cn}



\author[2,3]{Bo Yang}[orcid=0000-0003-1927-8419]      
\ead{ybo@jlu.edu.cn}
\credit{Conceptualization, Writing - review \&
editing, Supervision}

\address[2]{School of Computer Science and Technology, Jilin University, Changchun, Jilin, 130012, China}

\address[3]{Key Laboratory of Symbolic Computation and Knowledge Engineering of Ministry of Education, Jilin University, Changchun 130012, China}

\cortext[1]{Corresponding author}



\begin{abstract}
Graph learning is a prevalent field that operates on ubiquitous graph data. Effective graph learning methods can extract valuable information from graphs. However, these methods are non-robust and affected by missing attributes in graphs, resulting in sub-optimal outcomes. This has led to the emergence of incomplete graph learning, which aims to process and learn from incomplete graphs to achieve more accurate and representative results. In this paper, we conducted a comprehensive review of the literature on incomplete graph learning. Initially, we categorize incomplete graphs and provide precise definitions of relevant concepts, terminologies, and techniques, thereby establishing a solid understanding for readers. Subsequently, we classify incomplete graph learning methods according to the types of incompleteness: (1) attribute-incomplete graph learning methods, (2) attribute-missing graph learning methods, and (3) hybrid-absent graph learning methods. By systematically classifying and summarizing incomplete graph learning methods, we highlight the commonalities and differences among existing approaches, aiding readers in selecting methods and laying the groundwork for further advancements. In addition, we summarize the datasets, incomplete processing modes, evaluation metrics, and application domains used by the current methods. Lastly, we discuss the current challenges and propose future directions for incomplete graph learning, with the aim of stimulating further innovations in this crucial field. To our knowledge, this is the first review dedicated to incomplete graph learning, aiming to offer valuable insights for researchers in related fields.\footnote{We developed an online resource to follow relevant research based on this review, available at https://github.com/cherry-a11y/Incomplete-graph-learning.git}
\end{abstract}



\begin{keywords}
Graph learning\sep Incomplete graphs\sep Incomplete graph learning\sep Attribute-missing graphs\sep Attribute-incomplete graphs\sep Robustness \sep
\end{keywords}

\maketitle


\section{Introduction}\label{Introduction}
Graphs, also known as networks, consist of nodes connected by edges and are ubiquitous in real-world applications. From in-depth analyses of social networks~\cite{DBLP:conf/www/LuCL24} to the design and optimization of transportation systems~\cite{DBLP:conf/nips/MariscaCA22, DBLP:journals/tnn/LvWWZXX24}, and the development of personalized recommendation systems~\cite{2020Handling, DBLP:journals/nn/FaroughiMJ25}, graphs play a pivotal role across these domains. Graph learning is a prominent field that aims to extract useful information from graph data. The core of graph learning lies in leveraging the graph structure to address various tasks, such as link prediction~\cite{DBLP:journals/tkde/ChenZCDX23}, node classification~\cite{DBLP:conf/ijcai/LiuZMDTWH23}, and graph classification~\cite{DBLP:journals/tnn/CuiBBWH24}. Using graph learning, we can uncover hidden patterns in graphs, such as community structures and node relationships, providing effective solutions to solve real-world problems. 

Early research in graph learning relied on traditional machine learning techniques, such as matrix factorization methods ~\cite{DBLP:conf/www/AhmedSNJS13}, but these methods are limited to processing small graphs~\cite{DBLP:conf/kdd/XuYFS07} or graphs with specific structures ~\cite{DBLP:conf/cscw/GuhaW15}. 
With the emergence of graph representation learning ~\cite{DBLP:journals/nn/JuFGLLQQSSXYYZWLZ24}, new opportunities have emerged in the field of graph learning. Graph representation learning enables the embedding of graph structures, including nodes and edges, into a low-dimensional space. This area encompasses two categories: random walk-based methods and graph neural network methods ~\cite{DBLP:journals/corr/abs-2212-08966}.
Random walk-based methods rely on capturing node proximity through contextual relationships within the graph structure. In contrast, graph neural networks (GNNs) learn node representations by aggregating and updating information from local neighborhoods. Compared to traditional and random walk-based methods, GNNs demonstrate significantly enhanced expressive capacity and superior performance, establishing them as the most prevalent methods in contemporary graph learning.

\noindent $\mathbf{Incomplete~graph~phenomenon}$
Although graph learning methods demonstrate considerable effectiveness in practical applications, they frequently assume that the attributes of graphs are complete, an assumption that does not always apply in real world situations~\cite{Chen_2022}. Using social networks as an example, some users, out of concern for their personal privacy, may selectively disclose partial information or refrain from sharing any personal information at all, leading to missing or incomplete user attribute information. Such incomplete graph data\footnote{It is important to note that the incomplete graphs referred to in this paper specifically pertain to those with incomplete attribute information. It is important to note that, besides incomplete attributes, incomplete graph structures also exist~\cite{DBLP:conf/cikm/0011XYCW21, graphstructurelearning}. However, it should be emphasized that graph structure learning, being a well-established research field, has developed a substantial theoretical foundation and key advances, which have been extensively reviewed and summarized in relevant review papers~\cite{zhu2022surveygraphstructurelearning}. Consequently, this review does not offer a detailed exploration of graph structure learning. Rather, the focus is specifically on learning from graphs with incomplete attributes, with the aim of providing targeted and insightful perspectives to researchers in this field.} poses a significant challenge to the training of existing graph learning models. 

\begin{figure*}[ht]
\centering
\includegraphics[width=0.9\columnwidth]{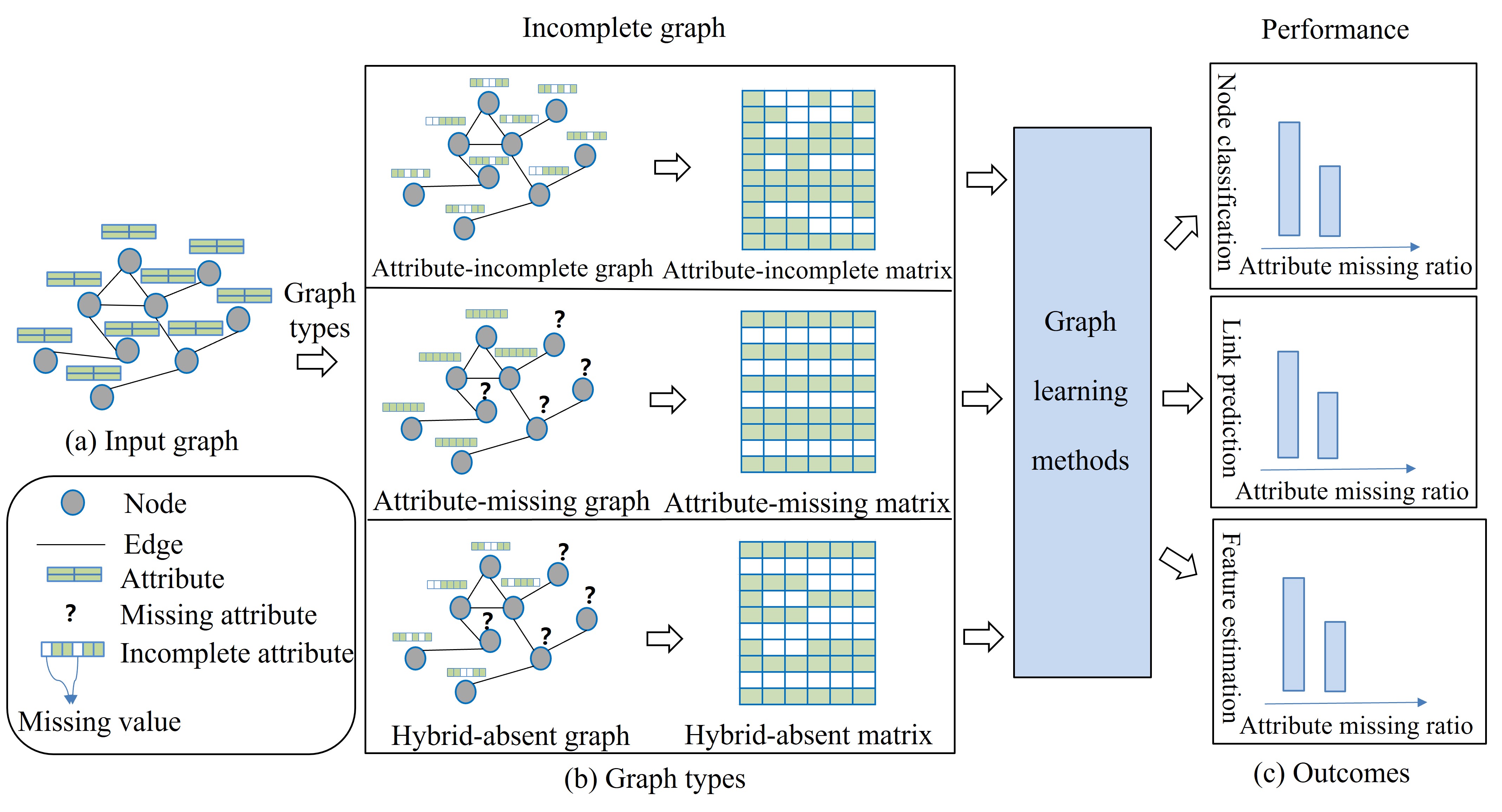}
\caption{Graph learning with incomplete graphs result in limit outcomes. }
\label{Fig1}
\end{figure*}

Due to incomplete attribute information, the models struggle to obtain sufficient data support~\cite{9724614}, preventing them from fully capturing the latent information and relationships, particularly in tasks that rely on attributes, thus significantly impairing the model's performance. More specifically, as illustrated in Figure \ref{Fig1}, when input graphs are provided, nodes within these graphs may lack crucial attribute information, leading to a deficiency of graph resources. This phenomenon is ubiquitous in the real world, as illustrated in Figure \ref{Fig1}(b), encompassing the cases of attribute-incomplete graphs, attribute-missing graphs, and hybrids of both (for detailed definitions, see Section \ref{section2.1.2}).
These attribute deficiencies can potentially limit the capabilities of graph learning models in feature extraction, inference, and prediction, ultimately leading to degraded model performance, as illustrated in Figure \ref{Fig1}(c). Therefore, the challenge of effectively learning from incomplete graphs has garnered increasing attention within the academic community, thereby further emphasizing the urgency and significance of addressing this issue.

\noindent $\mathbf{Incomplete~graph~learning.}$
To tackle the challenge of incomplete data phenomena, numerous methods have been introduced for independently and identically distributed (i.i.d.) data~\cite{DBLP:journals/pami/CaiHLN23, 10476707, DBLP:journals/pr/LiNWL23}, with image data~\cite{DBLP:journals/nn/QiuZZZX22} serving as a notable example. However, graph data fundamentally differ from these data types because its nodes are not i.i.d. but rather interconnected. Therefore, it is impractical to directly apply methods specifically designed for i.i.d. data to address incomplete graph data. 

\begin{figure*}[ht]
\centering
\includegraphics[width=0.7\columnwidth]{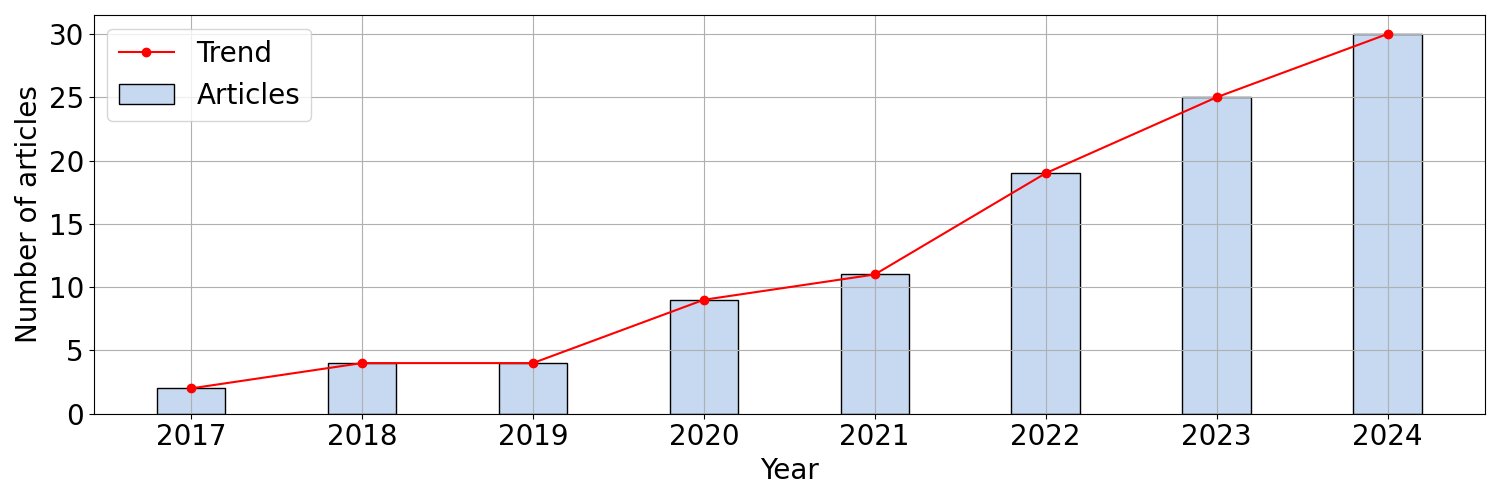}
\caption{Statistical analysis of published articles. }
\label{Fig2}
\end{figure*}

The significant impact of incomplete problems on the performance of graph learning has recently garnered considerable research attention~\cite{zhang2020inductivematrixcompletionbased, tu2023revisitinginitializingrefiningincomplete}, as shown in Figure \ref{Fig2}. The annual increase in the number of publications reflects the growing importance and increasing influence of tackling the challenge of incomplete graph learning. These research efforts aim to address various scenarios of practical applications (as shown in Section \ref{section6.4})~\cite{DBLP:conf/www/WangSZCH23, DBLP:conf/cvpr/XuBCCF23, 2020Handling, DBLP:conf/aaai/ChengZTG024} and can be categorized into several different types of incomplete graph learning methods (as shown in Section \ref{section3}-\ref{section5}). Each type possesses unique characteristics, necessitating the development of specialized techniques to effectively resolve the specific incompleteness issues encountered in each scenario. 
The fundamental principle of incomplete graph learning lies in recognizing that graph learning models often suffer from varying degrees of negative impacts on their overall performance when dealing with incomplete graphs. Consequently, a central objective of incomplete graph learning is to develop techniques and methodologies that can effectively address these attribute deficiencies, thus enhancing the robustness of graph learning models and improving their overall performance.


\noindent $\mathbf{Reasons~for~this~survey.}$
The motivation for conducting this comprehensive review derives from two key considerations. (1) \textbf{Absence of existing reviews.} Over the past decade, incomplete graph learning has emerged as a focal topic, attracting significant attention from the academic community. This field spans a broad range of research, from general methodologies to more specific applications like recommendation systems, traffic flow prediction, and anomaly detection. However, a comprehensive review of incomplete graph learning is still lacking. Therefore, this review aims to fill this gap by reviewing the key issues and methodologies associated with incomplete graphs.
(2) \textbf{Significance and contributions of this review.} This review aims to raise awareness within the graph machine learning community about the field of incomplete graph learning. With the widespread application of graph data across various domains, the incompleteness of graph data has become a critical issue that needs to be addressed. This review highlights the limitations and challenges of current methods, identifies key problems, introduces practical methods, and proposes future research directions to drive the advancement of this field. As the first comprehensive review exclusively dedicated to this topic, it fills a critical gap by providing an overarching perspective on incomplete graph learning. A comprehensive review not only offers a clearer understanding of the research trajectory but also provides valuable insights that can guide future research toward promising directions.

The main contributions of this review are threefold:
\begin{itemize}
\item
We present the first comprehensive review on incomplete graph learning, comprehensively covering the landscape of this field. Given the significance of this research area and the burgeoning number of publications, our review serves as a vital resource for both researchers and practitioners.
\item
In order to provide a comprehensive and structured overview of the field, we elucidate existing incomplete graph learning methods from the perspective of incomplete graph classification. The aim is to facilitate a deeper understanding of the existing literature and to clearly demonstrate commonalities and differences through structured classification. In addition, we discuss the practical aspects of incomplete graph learning.
\item
We identify potential future research directions in the field of incomplete graph learning, offering insights and guidance for those interested in driving the technologies in this rapidly evolving field.
\end{itemize}

The remainder of this review is organized as follows. In Section \ref{section2}, we categorize incomplete graphs into three types: (1) graphs with completely missing attributes (attribute-missing graphs), (2) graphs with partially missing attributes (attribute-incomplete graphs) and (3) hybrid-absent graphs. We provide definitions for these three types and introduce the relevant learning techniques related to incomplete graph learning. Section \ref{section3} presents learning methods for attribute-incomplete graphs, covering attribute imputation and label prediction approaches. Section \ref{section4} discusses learning methods for attribute-missing graphs, also covering attribute imputation and label prediction techniques. Section \ref{section5} focuses on learning methods for hybrid-absent graphs. Section \ref{section6} reviews the applications of incomplete graph learning, including dataset descriptions, incomplete strategies, evaluation metrics, and practical applications. Finally, Section \ref{section7} explores future research directions and concludes the review.

\section{Preliminaries}\label{section2}
In this section, we introduce the key terminologies and definitions related to graphs, incomplete graphs, and incomplete graph learning techniques.

\subsection{Taxonomy of graphs}\label{section2.1}
This section will introduce the fundamental types of graphs, including homogeneous graphs, heterogeneous graphs, bipartite graphs, as well as knowledge graphs. Subsequently, we summarize the types and definitions of incomplete graphs, serving as the foundation for incomplete graph learning.
\subsubsection{Taxonomy of the fundamental graphs}\label{section2.1.1}
\newtheorem{definition}{Definition}
\begin{definition}
A graph, denoted as $G = \{V, E, \mathbf{X}_v, \mathbf{X}_e, \phi, \varphi, T, R\}$, encapsulates a multitude of components. Specifically, $V$ denotes the set of nodes and $E$ denotes the set of edges that connect these nodes. The feature matrices $\mathbf{X}_v \in \mathbb{R}^{|V| \times d_v}$ and $\mathbf{X}_e \in \mathbb{R}^{|V| \times d_e}$ capture the attributes of nodes and edges, respectively, where $d_v$ and $d_e$ indicate the dimensionality of the feature spaces for nodes and edges. Furthermore, $T$ and $R$ represent node and edge types, respectively.
\end{definition}

For convenience, we introduce the notation $\mathbf{x}_v \in \mathbb{R}^{d_v}$ and $\mathbf{x}_e \in \mathbb{R}^{d_e}$ to represent the feature vectors of a node $v$ and an edge $e$, respectively. The functions $\phi: V \rightarrow T$ and $\varphi: E \rightarrow R$ map each node and edge to its respective type. This diversity in node and edge types results in the classification of graphs into different types. A homogeneous graph, characterized by having only one type of node ($|T| = 1$) and one type of edge ($|R| = 1$), ignores the distinction of different node and edge types, thereby simplifying the research process in graph analytics. In contrast, a heterogeneous graph includes multiple node and edge types ($|T| + |R| > 2$), enriching semantic content. Heterogeneous graphs can be further classified by the degree of their heterogeneity. A bipartite graph requires two distinct node types, with edges exclusively connecting nodes of different types, satisfying $|T| = 2$, $|R| = 1$, and $\phi(u) \neq \phi(v)$ for all $e\langle u,v \rangle \in E$. These bipartite graphs are extensively studied in recommender systems for modeling user-item interactions. A knowledge graph is a structured representation of facts, comprising nodes or entities, semantic descriptions, and relationships. The nodes can represent objects and concepts, while relationships denote connections between these nodes. The semantics, which encompass clearly defined types and properties, provide descriptions of the entities and their interrelations.

Since this paper focuses on incomplete graphs, we will not delve into the details of these fundamental graphs. For a comprehensive understanding of these graph types, we recommend consulting the relevant reviews cited.

\subsubsection{Taxonomy of the incomplete graphs}\label{section2.1.2}
\begin{figure*}[ht]
\centering
\includegraphics[width=1\columnwidth]{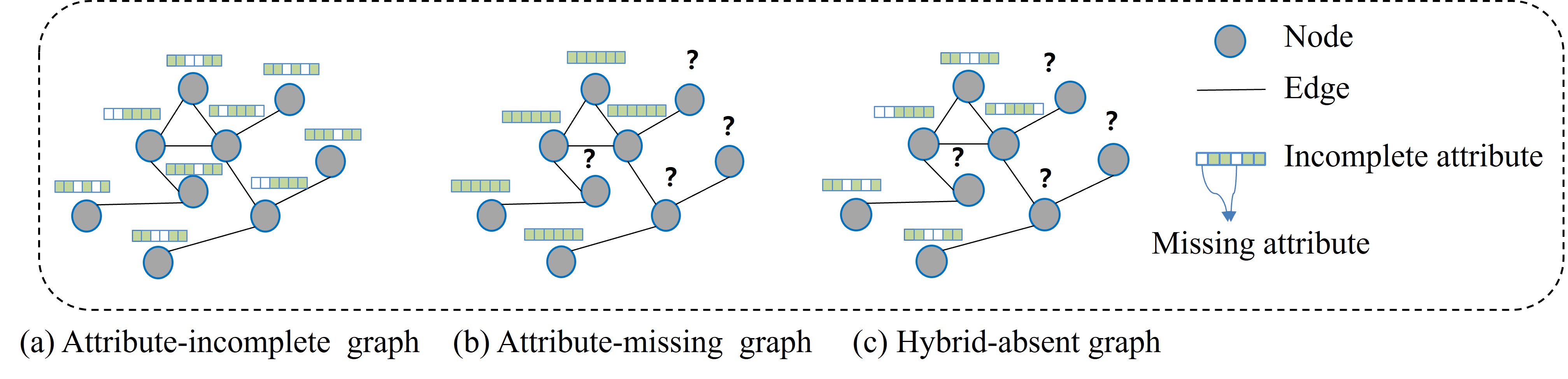}
\caption{Types of incomplete graphs. }
\label{Fig3}
\end{figure*}

Given a graph with attributes, we classify the graph into four types based on the completeness of the node attributes: (a) the attribute-complete graph; (b) the attribute-incomplete graph; (c) the attribute-missing graph; and (d) the hybrid-absent graph. The specific classification is shown in Table \ref{table1} and Figure \ref{Fig3}. Subsequently, we provide detailed definitions for each type of incomplete graph.

\begin{table}[]
\caption{Types and explanations of incomplete graphs. }
\label{table1}
\renewcommand{\arraystretch}{1.2}
\begin{tabular}{l|l|l}
\hline
& Incomplete graph types & Explanations \\ \hline
\multirow{4}{*}{Incomplete graphs} & Attribute-missing graph & \begin{tabular}[c]{@{}l@{}}Attributes of some nodes are completely\\ missing.\end{tabular} \\ \cline{2-3}
& Attribute-incomplete graph & \begin{tabular}[c]{@{}l@{}}Some attributes of some nodes are missing.\end{tabular} \\ \cline{2-3}
& Hybrid-absent graph & \begin{tabular}[c]{@{}l@{}}Both attribute-incomplete nodes and\\ attribute-missing nodes exist in the same graph.\end{tabular} \\ \hline
\end{tabular}
\end{table}

\begin{definition}
(Attribute-incomplete graph) A graph $\mathbf{G}  = (\textbf{V}, \textbf{X}, \textbf{A})$ is an attribute-incomplete graph that contains the node set $\textbf{V}=\left \{  {\textbf{v}_{1}, \dots, \textbf{v}_{N}} \right \}$, attribute matrix $\textbf{X}\in {R} ^{N\times D}$, and adjacent matrix $\textbf{A}\in \{0,1\} ^{N\times N}$. Attributes of certain dimensions in $\textbf{X}$ are missing. $N$ and $D$ denote the quantity of nodes and node attribute dimensions, respectively.
\end{definition}\begin{definition}
(Attribute-missing graph) A graph $\mathbf{G}  = (\textbf{V}, \textbf{X}^{o}, \textbf{A})$ is an attribute-missing graph that contains the node set $\textbf{V}=\left \{  {\textbf{v}_{1}, \dots, \textbf{v}_{N}} \right \}$, observed node attribute matrix $\textbf{X}^{o}\in {R} ^{N^{o}\times D}$, and adjacent matrix $\textbf{A}\in \{0,1\} ^{N\times N}$. Let $\textbf{X}^{m} \in {R} ^{N^{m}\times D}$ be a matrix of attribute-missing nodes, where $N^{m}$, $N$, $N^{o}$ and $D$ represent the quantity of attribute-missing nodes, nodes, attribute-observed nodes, and node attribute dimensions, respectively. $V^{o}$ and $V^{m}$ are the sets of attribute-observed and attribute-missing nodes, respectively. $N = N^{o} + N^{m}$, $V = V^{o}\cup V^{m}$, and $V^{o}\cap V^{m}=\emptyset$. $\mathbf{G}  = (\textbf{V}, \textbf{X}^{o}, \textbf{A})$ is an attribute-missing graph if $N^{o}$ is less than the number of graph nodes, i.e., $N^{o}<N$.
\end{definition}\begin{definition}
(Hybrid-absent graph) A graph $\mathbf{G}  = (\textbf{V}, \textbf{X}^{o}_{o}, \textbf{A})$ is a hybrid-absent graph consisting of the node set $\textbf{V}=\left \{  {\textbf{v}_{1}, \dots, \textbf{v}_{N}} \right \}$, observed node attribute matrix $\textbf{X}^{o}_{o}\in {R} ^{N^{o}\times D}$, and adjacent matrix $\textbf{A}\in \{0,1\} ^{N\times N}$. Attributes of certain dimensions in $\textbf{X}^{o}_{o}$ are missing. $N$ and $D$ represent the number of nodes and the dimensions of the attributes of the nodes, respectively. $\mathbf{G}  = (\textbf{V}, \textbf{X}^{o}_{o}, \textbf{A})$ is a hybrid-absent graph if $N^{o}$ is less than the number of graph nodes.
\end{definition}

\subsection{Relevant learning techniques}\label{section2.2}
This section will introduce the key techniques and methodologies utilized in incomplete graph learning, specifically encompassing data imputation learning, label prediction learning, and graph representation learning.

\subsubsection{Data imputation learning}\label{section2.2.1}
\begin{figure*}[ht]
\centering
\includegraphics[width=1\columnwidth]{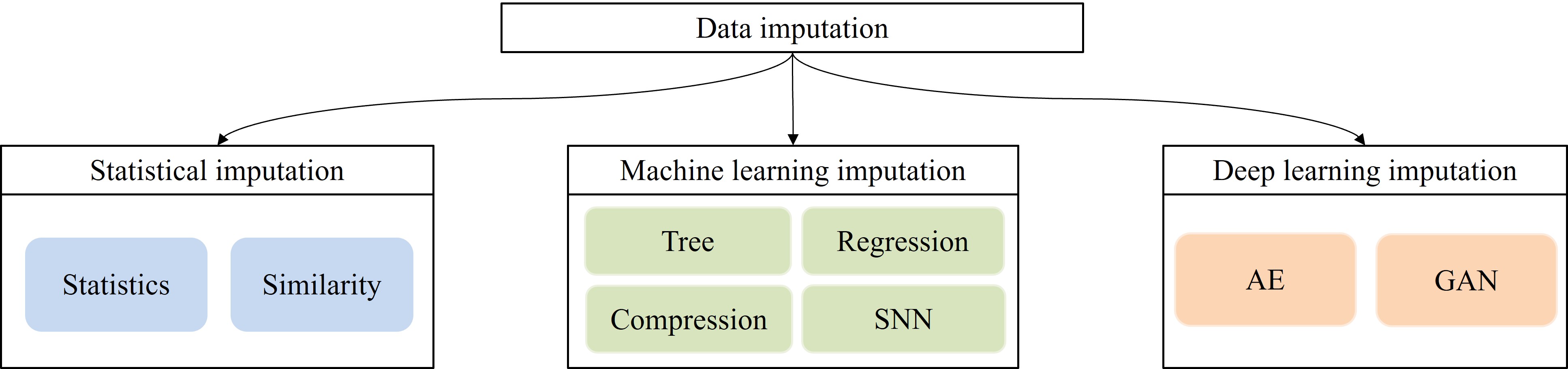}
\caption{Types of data imputation techniques. }
\label{Fig4}
\end{figure*}

Data imputation~\cite{2009Missing, 2020Missing} is an effective approach for tackling the problem of missing attributes. As a data processing technique, it is designed to estimate and fill in missing data, thereby resolving the problem of incomplete data and facilitating more efficient data analysis, machine learning modeling, and other data-related tasks. Data imputation techniques can be broadly categorized into three types (as shown in Figure \ref{Fig4}): statistical-based, machine learning-based, and deep learning-based methods~\cite{DBLP:journals/tkde/MiaoWCGY23}.

\begin{figure*}[ht]
\centering
\includegraphics[width=1\columnwidth]{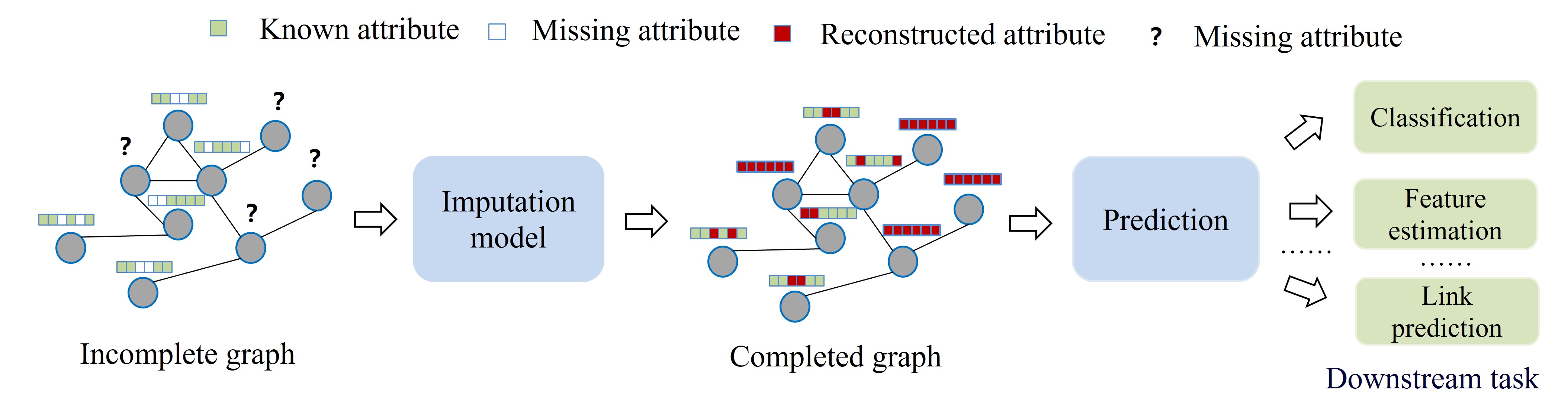}
\caption{Illustration of data imputation on graphs. }
\label{Fig5}
\end{figure*}

Statistical-based data imputation methods commonly utilized include statistical methods and similarity methods~\cite{DBLP:journals/tsmc/FarhangfarKP07, DBLP:conf/isese/TwalaCS05}. These approaches exploit statistical or similar information derived from the existing data to deduce the values of the missing data points. Machine learning-based data imputation methods utilize machine learning algorithms to learn the underlying patterns and correlations within the data, predicting the missing values. These methods include tree-based models~\cite{DBLP:journals/bioinformatics/StekhovenB12, DBLP:conf/kdd/ChenG16}, regression-based models~\cite{2011Multiple}, compression-based methods~\cite{DBLP:journals/adac/JossePH11}, and Shallow Neural Network-based models (SNN)~\cite{DBLP:conf/icml/MuzellecJBC20}. Deep learning-based data imputation methods utilize deep neural networks to impute missing data, such as those based on deep autoencoders (AE)~\cite{DBLP:conf/icml/MatteiF19, DBLP:journals/pr/NazabalOGV20} and those based on generative adversarial networks (GAN)~\cite{DBLP:journals/nn/SpinelliSU20, pmlr-v80-yoon18a}. The data imputation techniques mentioned above do not take into account the interactive relationships between nodes. In recent years, targeted methods have been proposed specifically for graph data, with the detailed process illustrated in Figure \ref{Fig5}. Among them, GNNs have gained widespread attention due to their unique ability to not only utilize sample information but also exploit the complex relationships among samples, rendering them a promising method for data imputation.

In general, data imputation techniques have made remarkable progress and a series of methods have been proposed. However, the current mainstream methods are mostly focused on the processing of Euclidean space data, and the imputation techniques for graph-structured data still need further research and development. Given that the main focus of this paper is not a comprehensive review of data imputation methods, it is recommended that researchers interested in this area refer to the relevant literature for more detailed information.

\subsubsection{Label prediction learning}\label{section2.2.2}
\begin{figure*}[ht]
\centering
\includegraphics[width=0.7\columnwidth]{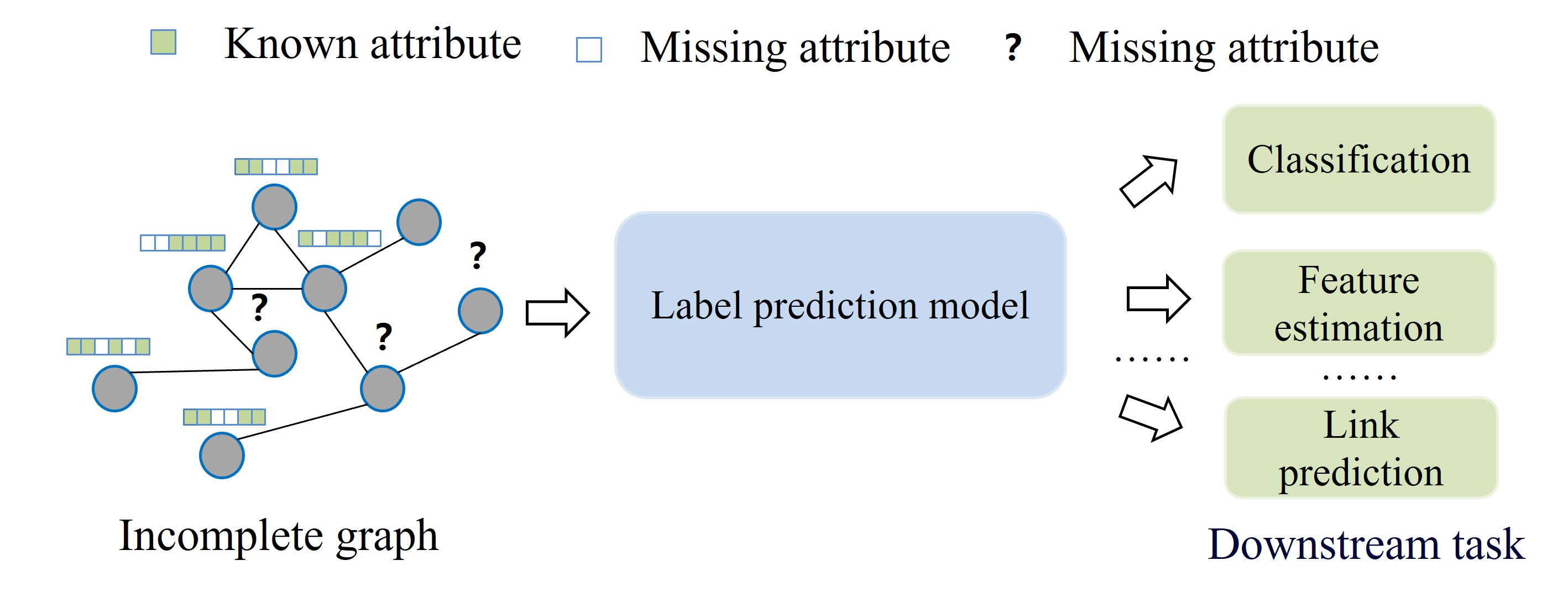}
\caption{Illustration of label prediction learning. }
\label{Fig6}
\end{figure*}

In data imputation methods, the missing attributes are first estimated and completed using imputation techniques, followed by the application of graph learning methods. However, the separate handling of feature imputation and graph learning can introduce noise and other issues, resulting in performance degradation and instability. Considering the limitations of data imputation methods, researchers have suggested an alternative technique centered on label prediction learning. This technique eliminates the need for imputation by utilizing a label-task-driven model that can directly learn from graphs containing missing attributes~\footnote{Data imputation methods focus on strategies to effectively complete missing attributes, while label prediction methods prioritize model adaptation to specific tasks and applications, guided by label information.}, as exemplified in Figure \ref{Fig6}. The mainstream approach in label prediction learning technology currently involves combining methods such as distribution learning~\cite{DBLP:journals/fgcs/TaguchiLM21}, feature propagation~\cite{DBLP:conf/icassp/LeiFWQHPY23}, and contrastive learning~\cite{zhang2022completingnetworkslearninglocal} with GNN to learn features or their distributions for predicting downstream tasks. In contrast to data imputation methods, existing label prediction methods have not yet been systematically classified, and specific methods are described in detail in Sections \ref{section3.2} and \ref{section4.2}.

Despite the immaturity of such methods, given the impact of noise and model complexity of data imputation methods, the development label-driven methods for handling graphs with missing attributes, represents a meaningful research direction. These approaches aim to circumvent the limitations of traditional data imputation techniques and directly address the challenges posed by missing attributes in graphs, thereby enhancing the robustness and effectiveness of graph-based models.

\subsubsection{Graph representation learning}\label{section2.2.3}
Graph representation learning~\cite{DBLP:journals/nn/JuFGLLQQSSXYYZWLZ24, DBLP:conf/kdd/JinMWLTCQTSYZY21} is a powerful and efficient approach to graph analysis. It offers an effective alternative to traditional graph engineering techniques by embedding graph structures (e.g., nodes and edges) into a low-dimensional space while preserving crucial structural information. Formally, let $\mathbf{h}_v \in \mathbb{R}^d$ denote the representation of node $v$. Node representation learning can be formulated as:
\begin{equation}\label{equ:1}
\begin{aligned}
\mathbf{h}_v = f(v, G; \theta_f),
\end{aligned}
\end{equation}
where $\theta_f$ represents the learnable parameters of the graph representation learning function $f(\cdot; \theta_f)$.

In recent years, GNNs~\cite{DBLP:journals/tnn/WuPCLZY21} have garnered increasing attention as an effective approach for graph representation learning.  
Many existing GNNs adopt the message-passing mechanism, which involves propagating and aggregating information based on the graph structure, such as GCN~\cite{Kipf2016SemiSupervisedCW}, GraphSAGE~\cite{DBLP:conf/nips/HamiltonYL17},SP-GNN~\cite{DBLP:journals/nn/ChenYHLPWZ23} etc. These GNNs consist of three functions: the message function, the attribute aggregation function, and the node attribute update function. The message passing process is:
\begin{equation}\label{equ:2}
\begin{aligned}
x_{v}^{l+1} = h_{l}(x_{v}^{l}, \varphi_{l}(\left \{ m_{l}(x_{v}^{l}, x_{u}^{l}, e_{v,u}) | u\in N(v)\right \} ) ),
\end{aligned}
\end{equation}
where $m_{l}$, $\varphi_{l}$ and $h_{l}$ are the message, attribute aggregation, and node attribute update functions, respectively. $x_{v}^{l}$ and $x_{u}^{l}$ are the representations of nodes $v$ and $u$. $N(v)$ denotes the neighbors of node $v$, $e_{v,u}$ denotes the edge weight of edge $\left \{v, u\right \}\in \textbf{E}$.

Furthermore, graph-level representations~\cite{DBLP:conf/nips/YingY0RHL18} are often necessary for graph-level tasks, such as molecular classification~\cite{DBLP:conf/ijcai/GuoGN0IMW0WZC23}. These representations typically leverage learned substructure representations (e.g., nodes~\cite{DBLP:conf/iclr/XuHLJ19} or edges~\cite{DBLP:conf/aaai/YuL0023}) for further aggregation to obtain the graph representation. Given a graph $G$ and considering node representations as the foundation, graph representation learning can be formulated as:
\begin{equation}\label{equ:2}
\begin{aligned}
\mathbf{h}_G = \text{READOUT}(\{\mathbf{h}_v: v \in V\}; \theta_r),
\end{aligned}
\end{equation}
where $\text{READOUT}(\cdot)$ denotes the readout function, tasked with aggregating node representations into a graph representation, it is parameterized by $\theta_r$. The $\text{READOUT}(\cdot)$ can be categorized into two approaches: global-pooling~\cite{Kipf2016SemiSupervisedCW} and hierarchical-pooling~\cite{DBLP:conf/iclr/ZhengH0KWS22}.

\subsection{Summary and discussion}\label{section2.3}
\begin{figure*}[ht]
\centering
\includegraphics[width=1\columnwidth]{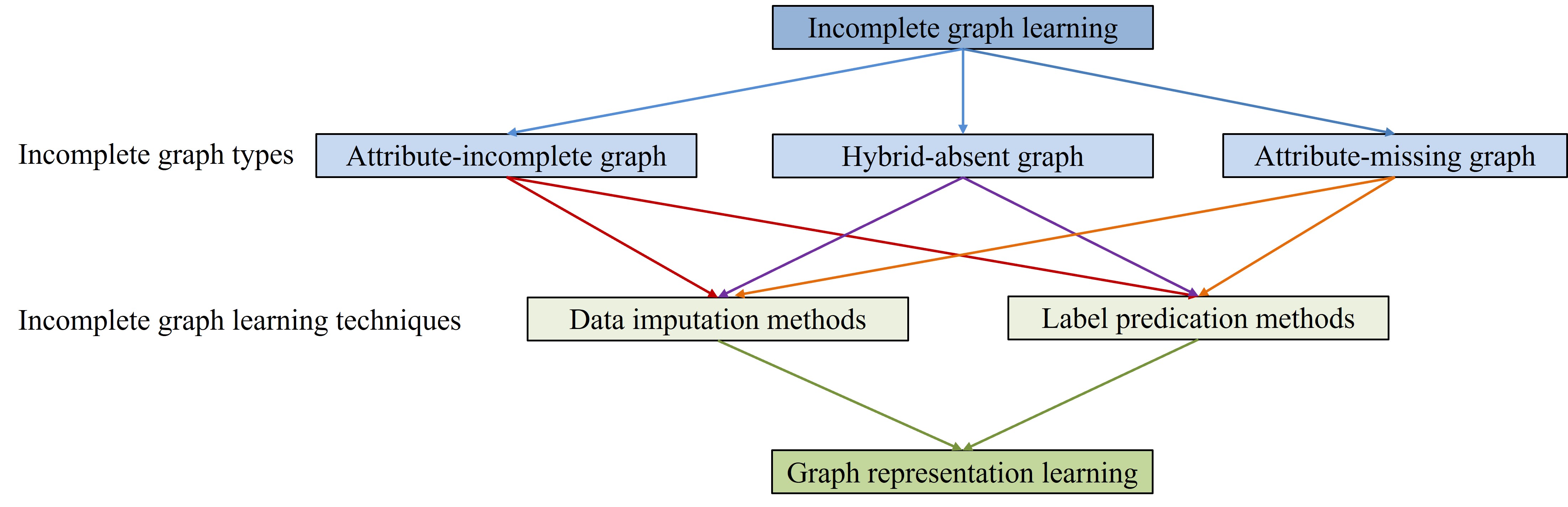}
\caption{Types of incomplete graphs and learning techniques. }
\label{Fig7}
\end{figure*}

In this section, we have conducted a detailed classification and in-depth comparative analysis of incomplete graphs, based on the types of attribute missingness, thereby establishing a solid foundation for the categorization of incomplete graph learning methods. According to the type of incomplete graphs, we categorize existing incomplete graph learning methods into three categories: (1) methods for attribute-incomplete graphs; (2) methods for attribute-missing graphs and (3) methods for hybrid-absent graphs. It is worth noting that data imputation learning, label prediction learning, and graph representation learning, as the current mainstream and promising technical approaches to addressing the challenges of incomplete graphs, hold significant importance. Therefore, this section systematically introduces the relevant techniques for incomplete graph learning, with the aim of facilitating a deep understanding of the internal mechanisms and advantages of these methods, thereby ensuring research continuity and depth. Figure \ref{Fig7} presents the types of incomplete graphs and the key techniques employed, along with the relationships between them. Specifically, incomplete graphs are categorized into three types: attribute-incomplete graph, attribute-missing graph, and hybrid-absent graph. All three types of incomplete graph learning methods employ the key techniques discussed in Section \ref{section2.2} to address missing attributes. Furthermore, both attribute imputation and label prediction learning methods leverage graph representation learning approaches.

\section{Attribute-incomplete graph learning methods}\label{section3}

In real-world applications, attribute-incomplete graphs are ubiquitous. Methods for handling such graph data can be classified into two categories. The first category focuses on completing the missing data through imputation. The second category includes label prediction methods, which are task-driven and train models directly on incomplete data without any imputation.

\subsection{Data imputation methods}\label{section3.1}
In fields such as complex networks and data science, the widespread issue of attribute missingness in graphs poses a significant challenge, often distorting data analysis and hindering its efficiency. Data imputation techniques are crucial for addressing this issue, as they impute the missing data in the input or latent space. To address attribute-incomplete learning problems, considerable efforts have been made to develop various imputation strategies on attribute-incomplete graphs. The common goal of these methods is to generate node features\footnote{In this paper, 'node features' and 'node attributes' are used interchangeably, both referring to the characteristics or properties of nodes.} or latent representations for nodes with missing attributes. In the context of incomplete graph learning tasks, we further categorize data imputation methods into traditional matrix completion methods and graph learning methods. 

\subsubsection{Traditional matrix completion methods}\label{section3.1.1}
Matrix completion, a powerful mathematical method, recovers missing values in the matrix, thereby preserving the completeness of the data. This process not only enhances the accuracy of completed matrix but also improves the reliability of subsequent data analysis, making it particularly effective for attribute-incomplete graphs. As a result, it has become an invaluable tool for a wide range of applications, including bioinformatics~\cite{DBLP:journals/jbcb/LiZLNHYL22}, social network analysis~\cite{DBLP:conf/lcn/MahindreJGP19}, bioinformatics~\cite{DBLP:journals/jbcb/LiZLNHYL22}, recommendation systems~\cite{DBLP:conf/ijcai/WangGD18}, and beyond. 

\begin{figure*}[ht]
	\centering
	\includegraphics[width=1\columnwidth]{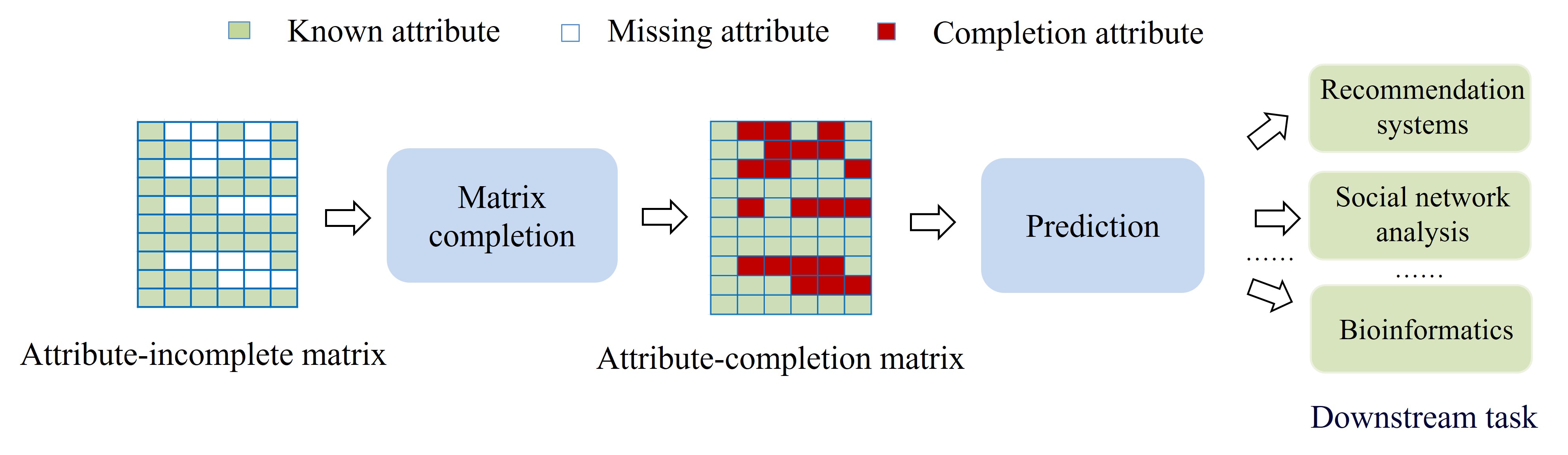}
	\caption{Illustration of matrix completion methods. }
	\label{Fig8}
\end{figure*}

When handling attribute-incomplete graphs, traditional matrix completion methods (as illustrated in Figure \ref{Fig8}) typically rely on the joint distribution of node attributes, neglecting the inherent graph structure, which is critical for precise imputation. For example, joint modeling methods, such as Bayesian strategies~\cite{2014Multiple}, matrix completion approaches~\cite{DBLP:journals/focm/CandesR09, DBLP:journals/jmlr/HastieMLZ15, DBLP:journals/tsp/ZhangTS21}, and generative adversarial networks~\cite{pmlr-v80-yoon18a, 2020GAMIN, DBLP:journals/nn/WangLLY21}, perform imputation by sampling from predictive distributions. An alternative joint modeling approach involves the iterative imputation of each variable's values using chained equations~\cite{2011MICE}, which are constructed based on other variables~\cite{2010Multiple, 0Flexible, DBLP:conf/icml/MuzellecJBC20}. Meanwhile, discriminative models such as random forests~\cite{DBLP:journals/pr/XiaZCLPYN17}, optimal transport-based distribution constraints~\cite{DBLP:conf/icml/MuzellecJBC20}, and causal-aware imputation~\cite{DBLP:conf/nips/KyonoZBS21} frequently rely on stringent assumptions, which can restrict their adaptability to address complex data types. Overall, the main limitation of traditional matrix completion methods is their inability to leverage the graph structural information present in graph-based scenarios.

Given our primary focus on incomplete graphs, we have refrained from an exhaustive exploration of traditional matrix completion techniques designed for Euclidean space data, as they are limited in their ability to leverage the inherent structural characteristics of graphs. Instead, we emphasize the importance of encouraging methods that motivate the development of approaches specifically tailored to graph-structured data, thereby yielding more accurate and reliable imputation results for incomplete graphs.


\subsubsection{Graph learning methods}\label{section3.1.2} 
Graph learning methods, which take advantage of both attribute and structural information (as illustrated in Figure \ref{Fig5}), have become effective data imputation strategies. Recent scholarly work has highlighted the potential of graph learning models in addressing imputation tasks. You et al.~\cite{2020Handling} introduced GRAPE, a framework that redefines the data input task as a bipartite graph link prediction learning process, then leverages GNNs to obtain better graph representations. Wei et al.~\cite{DBLP:journals/ida/WeiPHHYLZ19} presented IANRW, a novel approach that integrates random walks to effectively combine topological and attribute information, thereby demonstrating strong adaptability in addressing missing data challenges. Spinelli et al.~\cite{DBLP:journals/nn/SpinelliSU20} introduced GINN, a graph imputation method that utilizes adversarial training with Graph Auto-Encoders (GAEs) to achieve reliable imputation results. Morales-Alvarez et al.~\cite{DBLP:conf/nips/Morales-Alvarez22} proposed VISL, a scalable methodology that integrates structure learning with deep learning to infer variable relationships and impute missing values, which makes it particularly effective for large-scale data. Gao et al.~\cite{DBLP:conf/aaai/GaoNCTLXZTL23} introduced the Max-Entropy Graph AutoEncoder (MEGAE), which addresses the issue of spectral collapse in GAEs by preserving the full spectrum of spectral components, thereby improving imputation accuracy. Moreover, Zhong et al.~\cite{DBLP:conf/aaai/ZhongGY23} introduced the Iterative Graph Generation and Reconstruction Model (IGRM), a framework that leverages 'friend networks' and end-to-end reconstruction for iterative refinement, thereby enhancing imputation accuracy through differentiated message passing. These advances illustrate the dynamic evolution of graph-based imputation techniques, each designed to address distinct challenges and enhance the quality of imputed data. Most of these methods employ GNNs to learn node representations and utilize attribute reconstruction or link prediction constraints to derive more robust representations. Commonly used reconstruction constraints are as follows:
\begin{equation}\label{equ:2}
\begin{aligned}
\mathcal{L}_{edge} =\sum \left \| \mathbf{E} -\mathbf{\hat{E}}  \right \|_{F}^{2},
\end{aligned}
\end{equation}
\begin{equation}\label{equ:2}
\begin{aligned}
\mathcal{L}_{att} =\sum \left \| \mathbf{X} -\mathbf{\hat{X}}  \right \|_{X}^{2},
\end{aligned}
\end{equation}
where $\mathbf{E}$ and $\mathbf{X}$ represent the sets of edges and attributes, respectively, while $\mathbf{\hat{E}}$ and $\mathbf{\hat{X}}$ denote the reconstructed sets of edges and attributes.

Beyond the graph learning methods mentioned previously, most current approaches are focused on recommendation systems. Early works, such as GC-MC~\cite{2017Graph}, IGMC~\cite{zhang2020inductivematrixcompletionbased}, and GRAPE~\cite{2020Handling}, laid the foundation by constructing bipartite interaction graphs\footnote{The construction of the bipartite graph is illustrated in Figure \ref{Fig13}}. These graphs use the adjacency matrix as additional data, improving imputation accuracy and demonstrating the potential of graph-based techniques for addressing incomplete data in recommendation tasks. Subsequently, GNN has emerged as a powerful method to predict missing attributes in attribute-incomplete graphs. Their ability to model complex graph structures and propagate information across nodes makes them particularly effective for this task. To address computational intricacies, the RGCNN framework~\cite{DBLP:conf/nips/MontiBB17} integrated multi-graph convolutional neural networks (GCNNs) with recurrent neural networks, enabling it to extract meaningful statistical patterns from the interactions between users and items. In addition, the model applies a learnable diffusion process to the rating matrix, which enables it to outperform pure GCN-based approaches in its original context. Yao et al.~\cite{2018Convolutional} introduced Convolutional Geometric Matrix Completion (CGMC), a graph-based recommendation method that employs a GCN approach with a novel design to model interactions between users and items. To enable generalization to unseen users/items during training and facilitate transfer learning to new tasks, IGMC~\cite{zhang2020inductivematrixcompletionbased} introduces an inductive graph-based matrix completion approach that leverages GNNs to learn local graph patterns related to ratings. Elmahdy et al.~\cite{2014Multiple} contributed an efficient and parameter-free matrix completion method that incorporates hierarchical graph side information. By integrating hierarchical graph clustering with iterative refinement of both the clustering process and matrix ratings, this method offers a novel approach to using graph structures for matrix completion. Furthermore, IMC-GAE~\cite{DBLP:conf/cikm/0005ZTZHD021}, an inductive matrix completion method utilizing GAEs, demonstrates the ability of GAEs to learn user and item-specific representations for personalized recommendations. By capturing local graph structures, IMC-GAE improves the performance of inductive matrix completion, further broadening the use of graph learning in recommendation systems. The general expression of the loss function for incomplete graph learning methods in recommender systems is as follows:
\begin{equation}
\begin{aligned}
\mathcal{L}_{recommender} =\frac{1}{\left \| (n_i,n_j)|\Omega _{n_i,n_j=1} \right \| }   \sum_{(n_i,n_j):\Omega _{n_i,n_j=1}}^{}F(r[n_i,n_j],r'[n_i,n_j]),
\end{aligned}
\end{equation}
where, $r[n_i,n_j]$ and $r'[n_i,n_j]$ are the true and predicted rating of $(n_i,n_j)$, respectively, while the $\Omega $ denotes a mask for unobserved ratings in the rating matrix. $F$ is the Mean Squared Error (MSE) or Cross Entropy (CE) function.
To summarize, graph learning methods for incomplete graphs have made significant advancements, with a particular emphasis on recommendation systems. From the development of bipartite graph models to advanced GNN-based approaches and hybrid techniques, these methodologies continue to push the limits of handling incomplete data, improving recommendation accuracy and personalization.

Furthermore, researchers have also proposed methods for other fields. The most common application is traffic prediction. For example, Wu et al.~\cite{DBLP:journals/iotj/WuXFW22} introduced the Multi-Attention Tensor Completion Network (MATCN), which constructs multidimensional representations in data with missing entries. MATCN reduces exposure bias by sparsely sampling historical fragments and using a gated diffusion convolution layer. It also incorporates spatial signal propagation and temporal self-attention modules, enabling effective representation aggregation and dynamic dependency extraction at the spatiotemporal level, thereby improving data completion in complex scenarios. Meanwhile, Kong et al.~\cite{DBLP:journals/kbs/KongZSZLY23} developed DGCRIN to simulate dynamic spatial relationships in road networks with incomplete traffic data. DGCRIN uses a graph generator that combines recurrently imputation and historical data to model temporal spatial correlations. Zhou et al.~\cite{DBLP:journals/corr/abs-2406-03511} proposed the Mask-Aware Graph Imputation Network (MagiNet), which uses an adaptive mask spatio-temporal encoder to extract latent representations from incomplete data, eliminating the need for pre-filling missing values. MagiNet also employs a spatio-temporal decoder with multiple blocks to capture spatial and temporal dependencies, effectively reducing over-smoothing during imputation.

\subsection{Label prediction methods}\label{section3.2}
Given the limitations of imputation methods, which tend to introduce noise and other issues, researchers have proposed an alternative approach based on label prediction (as shown in Figure \ref{Fig6}). This method leverages a label-task-driven model that directly learns from graphs with missing attributes.

\begin{figure*}[ht]
\centering
\includegraphics[width=0.7\columnwidth]{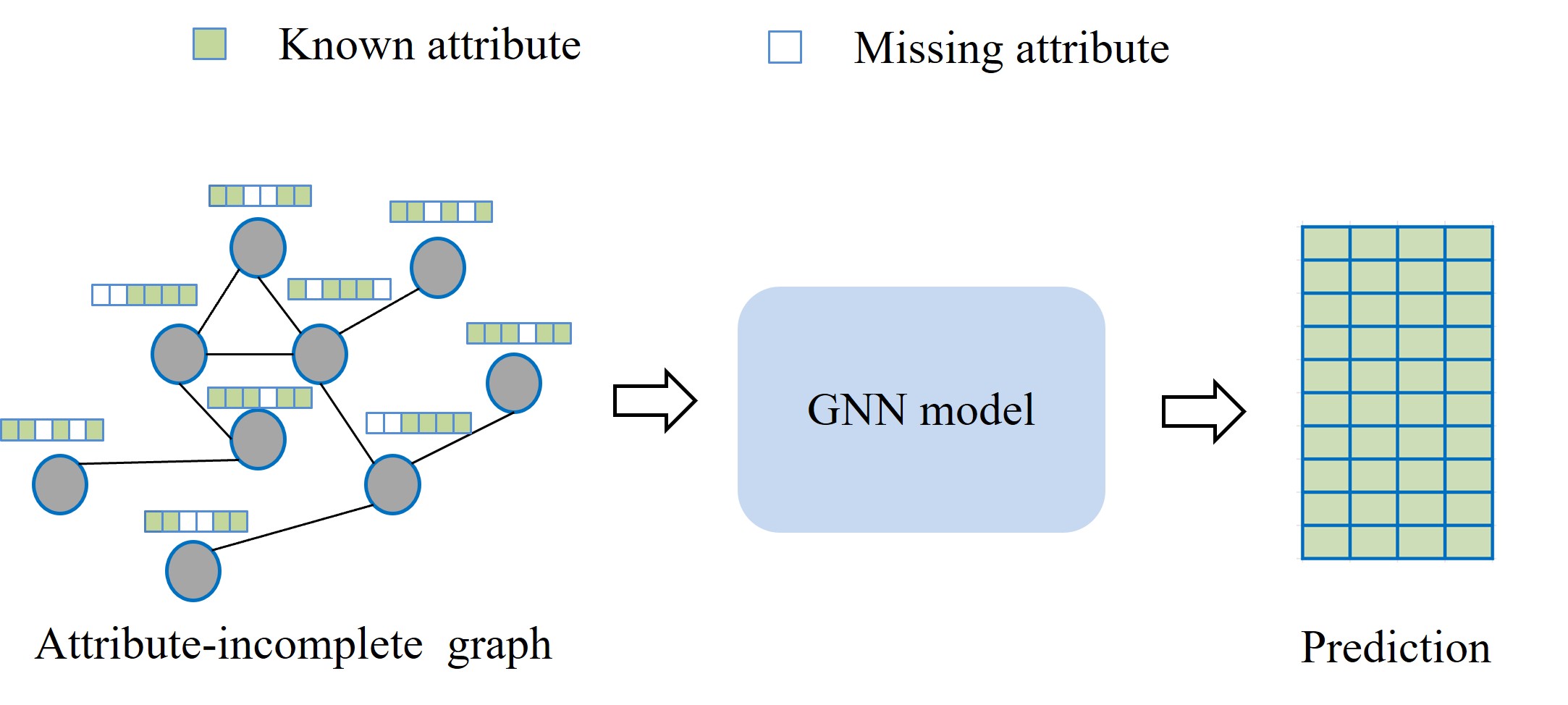}
\caption{Illustration of label prediction methods based on pure GNN for attribute-incomplete graphs. }
\label{Fig9}
\end{figure*}

For label prediction methods, Pure GNNs are an effective method on attribute-incomplete graphs. For example, Taguchi et al.~\cite{DBLP:journals/fgcs/TaguchiLM21} introduced a graph convolutional network without using imputation, called GCNmf. GCNmf integrates missing feature processing and graph learning into a single architecture. This model employs a Gaussian Mixture Model (GMM) to estimate missing data and simultaneously learns the parameters of the GMM and GNN. PaGNN~\cite{10495099} is a GCN-based model that uses a partial message-passing method to propagate observed attributes. Although these methods perform reasonably well when feature missing rates are low, they struggle in scenarios with high missing feature rates and do not scale effectively to large graphs. To avoid interference between features and structures, Jin et al.~\cite{DBLP:conf/aaai/Huo0LHYW23} designed T2-GNN, a framework that incorporates separate feature-level and structure-level teacher models. This framework provides targeted guidance to the student model through distillation. Figure \ref{Fig9} shows the generalized framework diagram of the pure GNN method for attribute-incomplete graphs.

\begin{figure*}[ht]
\centering
\includegraphics[width=1\columnwidth]{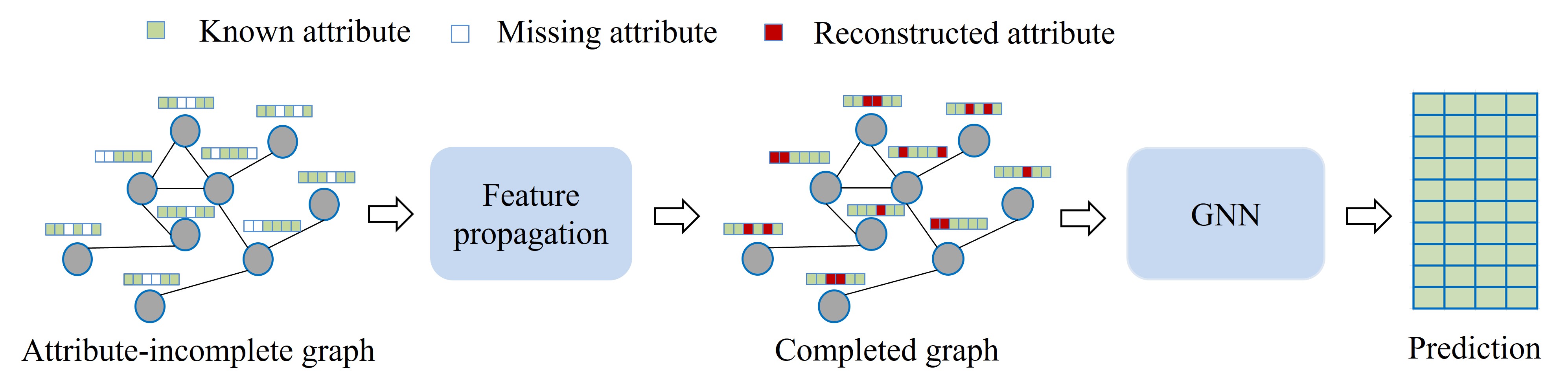}
\caption{Illustration of label prediction methods based on feature propagation for attribute-incomplete graphs. }
\label{Fig10}
\end{figure*}

Some researchers have proposed methods based on feature propagation (FP)~\cite{DBLP:conf/log/RossiK0C0B22}. Figure \ref{Fig10} shows the generalized framework of the feature propagation method. Specifically, FP is a method designed to address scenarios with high missing attribute rates. FP iteratively propagates known attributes to nodes with missing attributes to reconstruct the unknown attributes. Then, the graph and reconstructed attributes are fed into a GNN model, which produces predictions. However, FP only considers pairwise relationships between data and assumes each node has equal influence on all neighbors, which cannot accurately capture the local geometric distribution of the data. To address this issue, Lei et al.~\cite{DBLP:conf/icassp/LeiFWQHPY23} proposed SGHFP, which constructs feature and pseudo-label hypergraphs to capture the local geometric distribution of the data. The fused hypergraph is then applied to a feature propagation model to reconstruct missing features. They also reconstruct the missing features through imputation, which minimizes Dirichlet energy:
\begin{equation}
\begin{aligned}
l(x,G)= \frac{1}{2}x^{T}\Delta x= \frac{1}{2} {\textstyle \sum_{ij}^{}} \theta _{ij}(x_{i}-x_{j})^{2} ,
\end{aligned}
\end{equation}
where $\theta _{ij}$ represents the individual elements of the normalized incidence matrix.

In addition, Pan et al.~\cite{DBLP:conf/nips/PanK21} proposed the contrastive learning-based method MCGC (Multi-view Contrastive Graph Clustering), which initially filters out undesirable high-frequency noise while preserving the geometric features of the graph by applying graph filtering, thus obtaining a smooth representation of the nodes. Moreover, MCGC obtains a consensus graph that is regularized by a graph contrastive loss. Yu et al.~\cite{DBLP:journals/corr/abs-2210-01803} proposed a federated learning-based approach, where the model employs shared embeddings for training the network, thereby avoiding the direct sharing of original data.

While the current development of label prediction methods is still in its infancy, their potential to better address downstream tasks renders them a promising direction for future research and application.
\begin{table*}[]
\caption{Comparison of characteristics for attribute-incomplete graph learning methods.} \label{table2}
\scriptsize
\begin{tabular}{llllp{2.8cm}}
\hline
\multicolumn{1}{l}{Methods} & \multicolumn{1}{l}{} & \begin{tabular}[c]{@{}l@{}}Main \ references\end{tabular} & Advantages & Disadvantage    \\ \hline
\multirow{16}{*}{\begin{tabular}[c]{@{}l@{}}Data imputation\\  methods\end{tabular}} & \begin{tabular}[c]{@{}l@{}}Traditional matrix \\ completion methods\end{tabular} & \begin{tabular}[c]{@{}l@{}}~\cite{2014Multiple}\\ ~\cite{DBLP:journals/focm/CandesR09}\\ ~\cite{ DBLP:journals/jmlr/HastieMLZ15}\\ ~\cite{DBLP:journals/tsp/ZhangTS21}\\ ~\cite{ pmlr-v80-yoon18a}\\ ~\cite{2020GAMIN}\\ ~\cite{2011MICE}\\ ~\cite{2010Multiple}\\ ~\cite{0Flexible} \\ ~\cite{DBLP:conf/icml/MuzellecJBC20}\\ ~\cite{DBLP:journals/pr/XiaZCLPYN17}\\ ~\cite{DBLP:conf/icml/MuzellecJBC20}\\ ~\cite{DBLP:conf/nips/KyonoZBS21} \end{tabular}      & \begin{tabular}[c]{@{}l@{}}No graph structure \\ information is required.\end{tabular}                                                                          & \begin{tabular}[c]{@{}l@{}}The graph structure \\information is not \\exploited;\\ Easy to introduce  noise.\end{tabular} \\ \cline{2-5}
& \begin{tabular}[c]{@{}l@{}}Graph learning \\ methods\end{tabular}                & \begin{tabular}[c]{@{}l@{}}~\cite{2020Handling}\\ ~\cite{DBLP:journals/ida/WeiPHHYLZ19}\\ ~\cite{DBLP:journals/nn/SpinelliSU20}\\ ~\cite{DBLP:conf/nips/Morales-Alvarez22}\\ ~\cite{DBLP:conf/aaai/GaoNCTLXZTL23} \\ ~\cite{DBLP:conf/aaai/ZhongGY23}\\ ~\cite{2017Graph}\\ ~\cite{zhang2020inductivematrixcompletionbased}\\ ~\cite{2020Handling}  \\ ~\cite{DBLP:conf/nips/MontiBB17}\\ ~\cite{2018Convolutional}\\ ~\cite{zhang2020inductivematrixcompletionbased}\\ ~\cite{2014Multiple}\\ ~\cite{DBLP:conf/cikm/0005ZTZHD021}\end{tabular}      & \begin{tabular}[c]{@{}l@{}}Both attribute and \\ structural information\\  can be exploited.\end{tabular}                                                       & \begin{tabular}[c]{@{}l@{}}Easy to introduce  noise.\end{tabular}                                                           \\ \hline
\multicolumn{2}{l}{\begin{tabular}[c]{@{}l@{}}Label prediction\\  methods\end{tabular}}                                                                                & \begin{tabular}[c]{@{}l@{}}~\cite{DBLP:journals/fgcs/TaguchiLM21}\\ ~\cite{10495099}\\ ~\cite{DBLP:conf/aaai/Huo0LHYW23}\\ ~\cite{DBLP:conf/log/RossiK0C0B22}\\ ~\cite{DBLP:conf/icassp/LeiFWQHPY23}\\ ~\cite{DBLP:conf/nips/PanK21}\\ ~\cite{DBLP:journals/corr/abs-2210-01803} \end{tabular}      & \begin{tabular}[c]{@{}l@{}}Both attribute and \\ structural information\\  can be exploited;\\ No external information\\  needs to be brought in.\end{tabular} & \begin{tabular}[c]{@{}l@{}}Missing attributes may not \\be estimated.\end{tabular}                       \\ \hline
\end{tabular}
\end{table*}

\subsection{Summary and discussion}\label{section3.3}
In the comprehensive analysis presented in Table \ref{table2}, we perform a detailed comparison of graph learning methods for attribute-incomplete graphs. In general, data imputation techniques occupy a central position among these learning approaches, highlighting their importance in addressing the challenge of attribute-incomplete graph learning. Notably, within data imputation methods, graph learning approaches particularly those leveraging GNNs have demonstrated remarkable performance. This not only validates the unique advantages of GNNs in handling incomplete graphs but also suggests their broad application potential in this context. In contrast, research on label prediction methods remains relatively scarce. However, with the emergence of end-to-end GNN technologies, this field is poised for new developmental opportunities. End-to-end GNNs have provided new insights for label prediction tasks, significantly accelerating the development of related methods and highlighting their substantial potential in future research and applications.

In conclusion, it must be acknowledged that attribute-incomplete graph learning methods still face many challenges. For instance, data imputation methods, while imputing the attributes, tend to introduce noisy information, indicating that the field is still evolving. Consequently, developing efficient and precise methods for attribute-incomplete graph learning remains a key challenge, requiring sustained research attention from the scientific community.

\section{Attribute-missing graph learning methods}\label{section4}
Compared to attribute-incomplete graphs, attribute-missing graphs pose a more significant challenge, as certain nodes lack attributes entirely. In recent years, attribute-missing graphs have attracted considerable attention, leading to the development of several effective methods.

This subsection will introduce attribute-missing graph learning methods, including data imputation methods and label prediction methods. Data imputation methods can be categorized based on the type of graph, with separate approaches for homogeneous and heterogeneous graphs. Due to the limited progress in label prediction methods, we will not provide a further classification of label prediction methods. 

\subsection{Data imputation methods}\label{section4.1}
The mainstream data imputation methods for attribute-missing graphs target both homogeneous and heterogeneous graphs. Therefore, in this section, we will introduce data imputation methods for homogeneous and heterogeneous graphs. Table \ref{table3} lists the summary of data imputation methods for attribute-missing graphs, and each method is briefly described below.
\begin{table*}[t]
\renewcommand{\arraystretch}{1.0}
\setlength\tabcolsep{8pt}
\caption{A summary of data imputation methods for attribute-missing graphs.}
\label{table3}
\scriptsize
\begin{tabular}{lp{2.2cm}p{2cm}p{2cm}p{2.0cm}p{0.3cm}l}
\hline
Graph types                                                                & Methods   & Key Component    & Applications    & Datasets & Year & Venue      \\ \hline
\multirow{35}{*}{\begin{tabular}[c]{@{}c@{}}Homogeneous \\ graph\end{tabular}} & \begin{tabular}[c]{@{}p{2cm}@{}}\raggedright SAT~\cite{Chen_2022}\end{tabular}
 & \begin{tabular}[c]{@{}l@{}}Graph autoencoders,\\ Adversarial \\ distribution matching\end{tabular} & \begin{tabular}[c]{@{}l@{}}Node classification,\\ Link prediction,\\ Attribute completion\end{tabular} & \begin{tabular}[c]{@{}l@{}}Cora,\\  Citeseer,\\ Pubmed,\\ Amazon-Computer,\\ Steam,\\ Coauther-CS, \\Amazon-Photo \end{tabular}           & 2020 & \begin{tabular}[c]{@{}l@{}}IEEE\\ Transactions\\ on Pattern \\Analysis and\\ Machine \\Intelligence\end{tabular}                                                                 \\ \cline{2-7}
& \begin{tabular}[c]{@{}p{2cm}@{}}\raggedright Amer~\cite{9765782}\end{tabular}        & \begin{tabular}[c]{@{}l@{}} \multirow{1}{*}{Generative} \\  \multirow{1}{*}{adversarial network,}\\  \multirow{1}{*}{Graph autoencoders,}\\ \multirow{1}{*}{Mutual information} \\  \multirow{1}{*}{constraint}\end{tabular} & \begin{tabular}[c]{@{}l@{}} \multirow{2}{*}{Node classification,}\\ \multirow{2}{*}{Link prediction,}\\ \multirow{2}{*}{Node clustering}\end{tabular}      & \begin{tabular}[c]{@{}l@{}} \multirow{1}{*}{Cora,}\\  \multirow{1}{*}{Citeseer,}\\  \multirow{1}{*}{Pubmed,}\\  \multirow{1}{*}{Amazon-Computer,}\\  \multirow{1}{*}{Coauther-CS,} \\  \multirow{1}{*}{Amazon-Photo} \end{tabular}                     &  \multirow{2}{*}{2022} & \begin{tabular}[c]{@{}l@{}} \multirow{2}{*}{IEEE Transactions} \\  \multirow{2}{*}{on Cybernetics}\end{tabular} \\ \cline{2-7}
& ITR~\cite{DBLP:conf/ijcai/TuZLLCZZC22}        & \begin{tabular}[c]{@{}l@{}} \multirow{3}{*}{GNN,}\\ \multirow{3}{*}{Data imputation}\end{tabular}                                                                     & \begin{tabular}[c]{@{}l@{}}\multirow{3}{*}{Attribute completion,}\\ \multirow{3}{*}{Node classification}\end{tabular}                   & \begin{tabular}[c]{@{}l@{}}\multirow{3}{*}{Cora,}\\ \multirow{3}{*}{Citeseer,}\\ \multirow{3}{*}{Amazon-Photo,}\\ \multirow{3}{*}{Amazon-Computer}\end{tabular}                                 & \multirow{3}{*}{2022} & \multirow{3}{*}{IJCAI}                                                                 \\ \cline{2-7}
& \begin{tabular}[c]{@{}p{2cm}@{}}\raggedright SVGA~\cite{yoo2023accuratenodefeatureestimation}\end{tabular}      & \begin{tabular}[c]{@{}l@{}}Variational \\ graph autoencoders\end{tabular}                                                           & \begin{tabular}[c]{@{}l@{}}Attribute completion,\\ Node classification\end{tabular}                   & \begin{tabular}[c]{@{}l@{}}Cora,\\  Citeseer,\\  Pubmed,\\ Amazon-Photo,\\ Amazon-Computer,\\ Steam,\\ Coauthor-CS,\\ Arxive\end{tabular} & 2023 & KDD                                                                   \\ \cline{2-7}
& \begin{tabular}[c]{@{}p{2cm}@{}}\raggedright AmGCL~\cite{zhang2023amgclfeatureimputationattribute}\end{tabular}     & \begin{tabular}[c]{@{}l@{}}Graph contrastive \\ learning,\\ Dirichlet energy \\ minimisation\end{tabular}                          & \begin{tabular}[c]{@{}l@{}}Attribute completion,\\ Node classification\end{tabular}                   & \begin{tabular}[c]{@{}l@{}}Cora,\\  Citeseer,\\  Pubmed,\\ Amazon-Computer, \\ Amazon-Photo,\\ Steam,\\ Coauther-CS,\end{tabular}           & 2023 & ArXive                                                                \\ \cline{2-7}
& \begin{tabular}{@{} l @{}}\raggedright \parbox{2.2cm} 
        { AIAE~\cite{XIA2024111583}} 
\end{tabular}         & \begin{tabular}[c]{@{}l@{}}\multirow{1}{*}{Graph autoencoders,}\\ \multirow{1}{*}{Mask,}\\ \multirow{1}{*}{Multi-scale}\end{tabular}                                                  & \begin{tabular}[c]{@{}l@{}}\multirow{1}{*}{Attribute completion,}\\ \multirow{1}{*}{Node classification}\end{tabular}                   & \begin{tabular}[c]{@{}l@{}}\multirow{1}{*}{Cora,}\\ \multirow{1}{*}{Citeseer,}\\ \multirow{1}{*}{Amazon-Photo,}\\ \multirow{1}{*}{Amazon-Computer}\end{tabular}                                & \multirow{1}{*}{2024} &  \begin{tabular}[c]{@{}l@{}}\multirow{1}{*}{Knowledge-Based}\\\multirow{1}{*}{Systems}\end{tabular}                                                              \\ \cline{2-7}
& \begin{tabular}[c]{@{}p{2cm}@{}}\parbox{2.2cm} \raggedright 
FairAC~\cite{guo2023fairattributecompletiongraph}\end{tabular}
     & \begin{tabular}[c]{@{}l@{}}\multirow{1}{*}{Autoencoders,}\\ \multirow{1}{*}{Data imputation,}\\ \multirow{1}{*}{Sensitive classifier}\end{tabular}                                     & \begin{tabular}[c]{@{}l@{}}\multirow{1}{*}{Node classification,}\\\multirow{1}{*}{Fairness task}\end{tabular}                          & \begin{tabular}[c]{@{}l@{}}\multirow{1.5}{*}{NBA,}\\ \multirow{1.5}{*}{Pokec-z and Pokec-n}\end{tabular}                                         & \multirow{1}{*}{2023} & \multirow{1}{*}{ICLR}                                                               \\ \hline
\multirow{20}{*}{\begin{tabular}[c]{@{}c@{}}Heterogeneous \\ graph\end{tabular}} & HGNN-AC~\cite{10.1145/3442381.3449914}   & \begin{tabular}[c]{@{}l@{}}\multirow{2}{*}{Attention mechanism,}\\ \multirow{2}{*}{Data imputation}\end{tabular}  & \begin{tabular}[c]{@{}l@{}}\multirow{2}{*}{Node classification}\end{tabular} & \begin{tabular}[c]{@{}l@{}} \multirow{2}{*}{DBLP,}\\ \multirow{2}{*}{ACM,}\\ \multirow{2}{*}{IMDB}\end{tabular}   &\multirow{2}{*}{2021} & \multirow{2}{*}{WWW}                                                                 \\ \cline{2-7}
&  HGCA~\cite{9724614}     & \begin{tabular}[c]{@{}l@{}}\multirow{2}{*}{Contrastive learning,}\\ \multirow{2}{*}{Data imputaion}\end{tabular}    
& \begin{tabular}[c]{@{}l@{}}\multirow{2}{*}{Node classification,}\\ \multirow{2}{*}{Node clustering}\end{tabular}                                                                              & \begin{tabular}[c]{@{}l@{}}\multirow{2}{*}{DBLP,}\\ \multirow{2}{*}{ACM,}\\ \multirow{2}{*}{Yelp}\end{tabular}                                                    & \multirow{2}{*}{2022} &  \begin{tabular}[c]{@{}l@{}}\multirow{2}{*}{IEEE} \\\multirow{2}{*}{Transactions on}\\\multirow{2}{*}{Neural Networks and}\\ \multirow{2}{*}{Learning Systems}\end{tabular}  \vspace{4pt}                                                                 \\ \cline{2-7}
&  HetReGAT-FC~\cite{LI2023424}    & \begin{tabular}[c]{@{}l@{}}\multirow{2}{*}{Residual graph} \\ \multirow{2}{*}{attention network,}\\ \multirow{2}{*}{Attention mechanism}\end{tabular}                                  & \begin{tabular}[c]{@{}l@{}}\multirow{2}{*}{Node classification,}\\ \multirow{2}{*}{Node clustering}\end{tabular}                        & \begin{tabular}[c]{@{}l@{}}\multirow{2}{*}{DBLP,}\\ \multirow{2}{*}{ACM,}\\ \multirow{2}{*}{IMDB}\end{tabular}                                                    & \multirow{2}{*}{2023} & \multirow{2}{*}{Information Sciences}                                                                   \\ \cline{2-7}
& AutoAC~\cite{zhu2023autoacautomatedattributecompletion}     & \begin{tabular}[c]{@{}l@{}}\multirow{2}{*}{Continuous}\\ \multirow{2}{*}{relaxation schema,}\\ \multirow{2}{*}{Differentiable} \\ \multirow{2}{*}{completion}\end{tabular}     & \begin{tabular}[c]{@{}l@{}}\multirow{2}{*}{Node classification,}\\ \multirow{2}{*}{Link prediction}\end{tabular}                        & \begin{tabular}[c]{@{}l@{}}\multirow{2}{*}{DBLP,}\\ \multirow{2}{*}{ACM,}\\ \multirow{2}{*}{IMDB,}\\ \multirow{2}{*}{LastFM}\end{tabular}                                           & \multirow{2}{*}{2023} & \multirow{2}{*}{ICDE}                                                                  \\ \cline{2-7}
&  RA-HGNN~\cite{ZHAO2024122945} & \begin{tabular}[c]{@{}l@{}}\multirow{2}{*}{Residual attention}\\ \multirow{2}{*}{mechanism,}\\ \multirow{2}{*}{Dropping some edges}\end{tabular}                                & \begin{tabular}[c]{@{}l@{}}\multirow{2}{*}{Node classification,}\\ \multirow{2}{*}{Node clustering}\end{tabular}                        & \begin{tabular}[c]{@{}l@{}}\multirow{2}{*}{DBLP,}\\ \multirow{2}{*}{ACM,}\\ \multirow{2}{*}{IMDB}
\end{tabular}                                                    & \multirow{2}{*}{2024} &  \begin{tabular}[c]{@{}l@{}}\multirow{2}{*}{Expert Systems}\\\multirow{2}{*}{with Applications}\end{tabular}   \\ \hline

       \end{tabular}
  
\end{table*}

\subsubsection{Data imputation methods on homogeneous graphs}\label{section4.1.1}
In attribute-missing graph learning, most existing methods focus on homogeneous graphs. To tackle the challenge of attribute completion in these graphs, researchers have developed a range of data imputation strategies. These methods leverage available node and edge information, combining advanced graph learning techniques and data imputation methods to impute missing attributes. The common goal of these methods is to generate features or latent representations for nodes that lack attributes.
 
Chen et al.~\cite{Chen_2022} introduced a novel problem in graph analysis in 2022, focusing on attribute-missing graphs, which differs from attribute-incomplete graphs. Furthermore, assuming a shared latent space across graphs, they proposed a GNN-based Structure-Attribute Transformer (SAT) algorithm based on distribution matching, aimed at imputing missing attributes. Jin et al.~\cite{9765782} proposed Amer, a game-theoretic GNN framework that combines attribute estimation with graph representation learning using game theory principles and mutual information optimization, thus forming a unified learning framework. Both SAT~\cite{Chen_2022} and Amer~\cite{9765782} employ generative adversarial strategies to tackle the problem of missing attributes. The generalized loss function of generative adversarial networks is as follows:
\begin{equation}
\begin{aligned}
\underset{G_s}{min}\underset{D_s}{max} V(D_s,G_s)=\underset{G_s}{min}\underset{D_s}{max}E_{x \sim p_{data} (x)}[logD_s(x)]+E_{z \sim p_{z} (z)}[log(1-D_s(G_s(z)))].
\end{aligned}
\end{equation}
The optimization of the adversarial loss $V(D_s, G_s)$ aims to (1) enable the generator $ G_s $ to produce realistic samples and (2) improve the ability of discriminator $D_s$ to distinguish real from generated samples.
Although these methods demonstrate potential in handling attribute-missing graphs, they still have some limitations. For instance, generative adversarial strategies may face challenges in capturing the intricate relationships between nodes, which makes it difficult to fully capture the intrinsic connections in the graph. Furthermore, the stringent distributional assumptions imposed on latent variables may limit the model's flexibility, thus weakening the discriminative ability and generalization performance of the learned representations.

\begin{figure*}[ht]
	\centering
	\includegraphics[width=1\columnwidth]{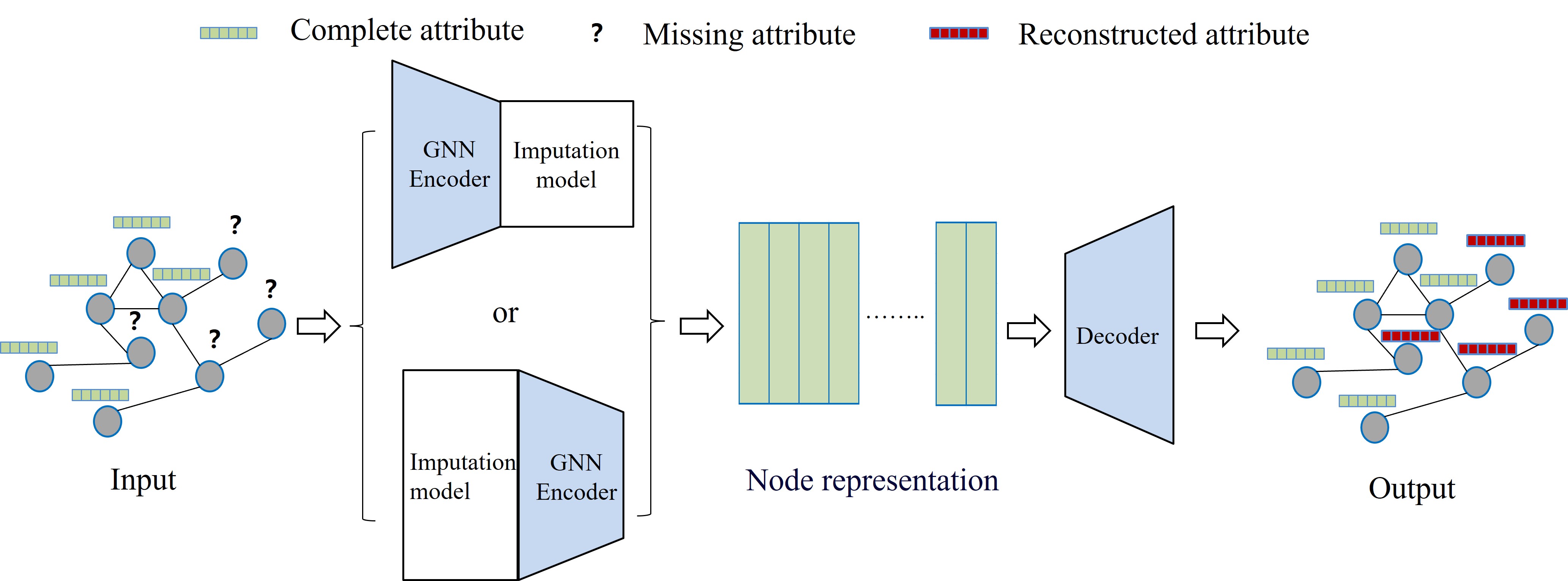}
	\caption{The general encoder-decoder framework of data imputation methods on homogeneous attribute-missing graphs.}
	\label{Fig11}
\end{figure*}

Based on the above analysis, Tu et al.~\cite{DBLP:conf/ijcai/TuZLLCZZC22} proposed a novel Initializing Then Refining (ITR) strategy, which combines reliable node attributes with structural information from the graph to generate effective representations for attribute-missing nodes. To preserve the validity of attribute-observed node representations and mitigate the impact of noise on these nodes, the ITR strategy employs a hybrid approach of attribute-observed and attribute-missing node representations to generate accurate representations for nodes in attribute-missing graphs. 
Yoo et al.~\cite{yoo2023accuratenodefeatureestimation} introduced the Structured Variational Graph Autoencoder (SVGA), which uses structured variational inference to impose constraints and regularization on the latent variable distribution. 
Zhang et al.~\cite{zhang2023amgclfeatureimputationattribute} proposed a method called Attribute-missing Graph Contrastive Learning (AmGCL). This method integrates node attribute completion and graph representation learning into a unified contrastive learning model. Notably, AmGCL introduces a novel combination of Dirichlet energy minimization and contrastive learning methods.
The methods discussed above almost all employ an encoder-decoder approach to address the problem of attribute-missing graphs. Figure \ref{Fig11} illustrates the general encoder-decoder framework that employs data imputation techniques. To address the limitations of graph autoencoder-based data imputation techniques in effectively integrating attribute and structural information during the encoding stage, as well as the inadequate design of decoder architectures, Xia et al.~\cite{XIA2024111583} proposed the Attribute Imputation AutoEncoder (AIAE) to handle attribute-missing graphs. Specifically, during the encoding phase, they employe a dual-encoder mechanism based on knowledge distillation, aiming to accurately encode both attribute and structural information into the representations of attribute-missing nodes. During the decoding stage, a multi-scale decoder incorporating a masking mechanism is introduced to enhance its expressive power, robustness, and generative capabilities. In response to the limitation of existing methods that focus solely on handling incomplete features or structures, Yuan et al.~\cite{DBLP:journals/corr/abs-2408-04845} proposed a Mutual Dual-Stream Graph Neural Network (MDS-GNN) that enables mutually beneficial learning between features and structures.

In addition to the previous research, some researchers have applied data imputation methods to particular application scenarios, such as single-cell analysis, multi-view clustering, and unfairness in graphs. 
In the field of single-cell analysis, Wen et al.~\cite{wen2022bichannel} introduced a novel imputation strategy that integrates spatial layout information of cells and the diversity across cell types to optimize the imputation process. Moreover, they introduced a 'mask-then-predict' framework aimed at modeling the imputation of missing data and further enhancing denoising performance, ultimately improving the accuracy and reliability of the analysis.
Multi-View Clustering (MVC)~\cite{DBLP:journals/kbs/WangYLF19, DBLP:journals/nn/ZhangPCLQ25} has gained significant attention to partition objects into different groups using multi-view features across various domains. Many methods have been proposed to address incomplete data in MVC~\cite{DBLP:journals/tsmc/WenZFZXZL23}, and recently, incomplete MVC methods on graphs have been introduced~\cite{DBLP:conf/ijcai/WangZLYZ19, DBLP:conf/aaai/GuoY19, DBLP:journals/tkde/LiangYX23, DBLP:journals/www/HeZCW23, DBLP:journals/tnn/CuiFHW24, DBLP:journals/tnn/LiCWL24, DBLP:journals/tnn/SunZ24}.
For example, Wen et al.~\cite{DBLP:journals/tmm/WenYZXWFZ21} proposed Adaptive Graph Completion-based Incomplete Multi-view Clustering (AGC IMC), which learns latent information from missing views while considering the information imbalance across different views.
To address the issue of partial missing views in some instances, Zhang et al.~\cite{ZHANG2022108412} proposed an effective method named IMNRL, which uses matrix factorization on multiple incomplete graphs to decompose them into a consensus non-negative representation and view-specific spectral representations, enabling simultaneous learning of both types of representations.
Yu et al.~\cite{DBLP:journals/corr/abs-2407-18170} proposed the Robust Incomplete Deep Attack (RIDA) model, which represents the first approach for robust gray-box poisoning attacks on incomplete graphs.
Moreover, Given the current limitations in addressing missing data in fair graph learning, Guo et al.~\cite{guo2023fairattributecompletiongraph} introduced FairAC, a dual optimization method that simultaneously improves attribute completeness and fairness. FairAC is the first approach to integrate graph attribute completion with fairness. Its core innovation lies in the application of an attention mechanism, which not only tackles attribute missingness but also reduces feature-level and topological-level unfairness, thereby enhancing both fairness and accuracy in graph learning. Furthermore, Cui et al.~\cite{cui2022positionalstructuralnodefeatures} investigated the use of artificial node features in applying GNNs to non-attributed graphs. They classify these features into two categories according to the information they assist GNNs in capturing, which are positional and structural node features.

In summary, numerous imputation methods have been proposed for homogeneous attribute-missing graphs. Among them, most methods revolve around GAE, which fully demonstrates the effectiveness and broad prospects of GAE in dealing with missing attributes. Moreover, specific methods tailored to the needs of different domains have emerged, highlighting the prevalence of missing attributes in graphs across various fields and providing forward-looking insights for domains that have yet to address this issue. Despite the numerous existing methods, there remains a significant research gap, highlighting substantial potential for future studies in this area.

\subsubsection{Data imputation methods on heterogeneous graphs}\label{section4.1.2}
In real-world graph data, heterogeneous graphs are widely applied. For example, social networks, recommendation systems, and knowledge graphs, which contain multiple types of edges and nodes, are all heterogeneous graphs. Due to privacy concerns and other factors, these graphs inevitably contain nodes with completely missing attributes. Based on this, some studies have extended the issue of missing attributes to the heterogeneous graphs.

In 2021, Jin et al.~\cite{10.1145/3442381.3449914} introduced the first Heterogeneous GNN model based on Attribute Completion (HGNN-AC). The model computes node topological embeddings and then uses an attention mechanism to complete missing node attributes by aggregating attributes from neighboring nodes. 
Due to challenges like data sparsity and cross-type information fusion, the HGNN-AC framework has limitations in fully integrating heterogeneous information. To address this, Zhao et al.~\cite{ZHAO2024122945} introduced the Heterogeneous Residual Graph Attention Network (RA-HGNN), which focuses on feature imputation. The core of RA-HGNN is its architecture, which leverages a residual graph attention mechanism to effectively explore topological features and improve the imputation of missing attributes, enhancing the model's ability to integrate heterogeneous information.
To address missing attributes and label scarcity in unsupervised learning, HGCA~\cite{9724614} introduces a contrastive learning strategy that simultaneously optimizes attribute completion and node representation within a heterogeneous framework. Additionally, HGCA improves completion accuracy by designing a network architecture that captures deep semantic relationships between nodes and attributes, enabling more precise imputation.
Existing attribute completion methods in heterogeneous graphs apply the same operation to all nodes. Zhu et al.~\cite{zhu2023autoacautomatedattributecompletion} recognized the semantic differences among nodes and proposed the AutoAC framework, a differentiable solution for fine-grained attribute completion. AutoAC introduces a comprehensive operation search space to cover diverse completion strategies. By applying a continuous relaxation technique, it transforms a non-differentiable search space into a differentiable one, enabling efficient optimization via gradient descent. Additionally, AutoAC models the completion process as a bi-level joint optimization task, improving both search efficiency and attribute completion accuracy.
To address the limitations of existing heterogeneous graph neural networks, which require pre-training and fail to fully utilize heterogeneous information, Li et al.~\cite{LI2023424} proposed the model named HetReGAT-FC. The approach first learns topological information with heterogeneous residual graph attention network, then applies an attention mechanism to complete missing features, and finally uses the completed graph to learn node representations.
Moreover, to address the issue in existing attribute completion methods where representation learning and attribute completion are treated separately, potentially leading to suboptimal results, as well as the problem of insufficient utilization of information from higher-order connected nodes during the attribute completion process, Chen et al.~\cite{DBLP:journals/kbs/ChenL24} proposed a method that integrates representation learning, attribute imputation, and heterogeneous graph learning into a unified model,Zhang et al.~\cite{zhang2025graph} presents a Bayesian framework and an expectation maximization (EM) algorithm aimed at jointly learning graph structures and recovering missing attributes from partially observed signals. We observe that many of the aforementioned methods employ graph attention mechanisms to learn the edge weights, as expressed by the following formula:
\begin{equation}
\begin{aligned}
c_{u,v} =softmax(\delta (T_{u}^{T}WT_{v})), 
\end{aligned}
\end{equation}
where, $(u, v)$ is node pair, $T_u$ and $T_v$ represent the representations of nodes $u$ and $v$, respectively. $W$ denotes the parametric matrix, and $\delta$ is the activation function.

\begin{figure*}[ht]
\centering
\includegraphics[width=1\columnwidth]{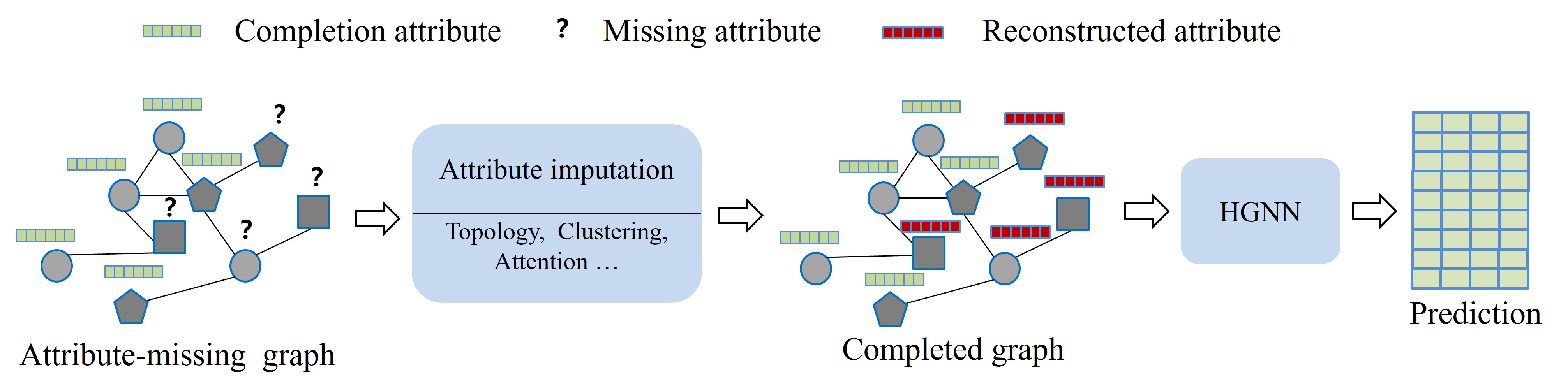}
\caption{The general framework of data imputation methods on heterogeneous attribute-missing graphs. }
\label{Fig12}
\end{figure*}
Figure \ref{Fig12} illustrates the comprehensive framework of data imputation methods for heterogeneous attribute-missing graphs. Addressing missing attributes in heterogeneous graphs is more challenging due to the greater variety of node and edge types. Although some imputation techniques for heterogeneous graphs have been proposed, the technology is still immature, and numerous issues remain. For example, how can we fully utilize the information in heterogeneous graphs to complete missing attributes? Furthermore, how can we integrate methods from homogeneous graphs to address the problem of missing attributes in heterogeneous graphs? Thus, data imputation methods tailored specifically for heterogeneous graphs represent a promising research direction.

\subsection{Label prediction methods}\label{section4.2}	
Methods for addressing attribute-missing graphs primarily utilize data imputation techniques, with some approaches based on label prediction techniques. These label prediction methods focus on specific application fields, such as event graph~\cite{wang2022schemaguidedeventgraphcompletion}, partial subgraph learning~\cite{kim2022modelsbenchmarksrepresentationlearning}, multi-view learning~\cite{9721669}, and community detection~\cite{10210121}. Table \ref{table4} lists the summary of label prediction methods for attribute-missing graphs, and each method is briefly described below.
\begin{table*}[H]
\renewcommand{\arraystretch}{1.2}
	\caption{A summary of label prediction methods for attribute-missing graphs.}
    \label{table4}
	\scriptsize 
    \begin{threeparttable}
	\begin{tabular}{p{2.5cm}p{2.8cm}llll}
		\hline
		Model     & Key Component                                                                                                                             & Applications                                                                                                           & Datasets                                                                                                                                    & Year & Venue  \\ \hline
		
	 \begin{tabular}{@{} l @{}}\raggedright \parbox{2.5cm} 
        {SchemaEGC~\cite{wang2022schemaguidedeventgraphcompletion}} 
      \end{tabular}                                        & \begin{tabular}[c]{@{}l@{}}Heuristic subgraph \\matching,\\ GNN\end{tabular}                                                                & Event graph completion                                                                                                 & \begin{tabular}[c]{@{}l@{}}Car-Bombings,\\  IED-Bombings, \\ Suicide-IED,\\  Pandemic,\end{tabular}                                             & 2022 & AKBC \\ \hline
	\begin{tabular}{@{} l @{}}\raggedright \parbox{2.5cm} 
        {C-GIN~\cite{zhang2022completingnetworkslearninglocal}} 
        \end{tabular}                                           &\multirow{1}{*} {Graph autoencoder}                                                                                                                        &\multirow{1}{*} {Link prediction}                                                                                                        & \begin{tabular}[c]{@{}l@{}}\multirow{1}{*}{Bio\_S,}\\ \multirow{1}{*}{Bio\_D,}\\ \multirow{1}{*}{Cora,}\\ \multirow{1}{*}{Co-Author}\end{tabular}                                                             & \multirow{1}{*}{2022} & \multirow{1}{*}{ArXive} \\ \hline
   
	PSI~\cite{kim2022modelsbenchmarksrepresentationlearning}       & InfoMax models                                                                                                                            & Partial subgraph learning                                                                                              & \begin{tabular}[c]{@{}l@{}}FNTN, \\ EM-User, \\ HPO-Metab\end{tabular}                                                                        & 2022 & CIKM   
         \\ \hline
		\begin{tabular}{@{} l @{}}\raggedright \parbox{2.5cm} 
        {DPMNE~\cite{wang2022deeppartialmultiplexnetwork}} 
\end{tabular}     & \begin{tabular}[c]{@{}l@{}}\multirow{1}{*}{Deep reconstruction loss,}\\ \multirow{1}{*}{Data consistency,}\\ \multirow{1}{*}{Proximity preservation}\end{tabular}                            & \begin{tabular}[c]{@{}l@{}}\multirow{1}{*}{Multiplex network}\multirow{1}{*} {learning,}\\ \multirow{1}{*}{Node classification}\end{tabular}                              & \begin{tabular}[c]{@{}l@{}}\multirow{1}{*}{DBLP,}\\ \multirow{1}{*}{Cora,}\\ \multirow{1}{*}{LastFM,}\\ \multirow{1}{*}{Flickr}\end{tabular}                                                                       & \multirow{1}{*}{2022} & \multirow{1}{*}{WWW}    \\ \hline
		LHGN~\cite{9721669}      & \begin{tabular}[c]{@{}l@{}}View-specific encoder\\  networks,\\ Heterogeneous graph \\ learning,\\ Aggregated representation\end{tabular} & Multi-view learning                                                                                                    & \begin{tabular}[c]{@{}l@{}}ORL, \\ YaleB, \\ PIE, \\ CUB, \\ Handwritten (HW), \\ Animal, \\ Caltech101-20, \\ LandUse-21, \\ Scene-15\end{tabular} & 2022 &  \begin{tabular}[c]{@{}l@{}}IEEE \\Transactions \\on \\Multimedia\end{tabular}    \\ \hline
    
		*~\cite{zhang2023learnabletopologicalfeaturesphylogenetic}     & Dirichlet energy                                                                                                                          & Phylogenetic inference                                                                                                 & DS1-DS8                                                                                                                                     & 2023 & ICLR   \\ \hline
		CSAT~\cite{10210121}     & \begin{tabular}[c]{@{}l@{}}Graph transformer,\\ Information transfer\\ mechanism,\\ Contrastive learning\end{tabular}                     & \begin{tabular}[c]{@{}l@{}}Node classification,\\ Link prediction,\\ Attribute completion,\\ Node clustering\end{tabular} & \begin{tabular}[c]{@{}l@{}}Cora,\\  Citeseer,\\  Pubmed,\\ Amazon-Photo,\\ Amazon-Computer,\\ Reddit,\\ Coauther-CS,\\ Yelp,\\ Flickr\end{tabular}                     & 2024 &  \begin{tabular}[c]{@{}l@{}} IEEE\\ Transactions\\on \\Computational\\ Social\\ Systems\end{tabular}  \\ \hline
	\begin{tabular}{@{} l @{}}\raggedright \parbox{2.5cm} 
        {ACTIVE~\cite{wang2022activeaugmentationfreegraphcontrastivelearning}} 
        \end{tabular}
    & \begin{tabular}[c]{@{}l@{}}Graph contrastive learning,\\ Graph consistency learning,\\ Autoencoders\end{tabular}                     & \begin{tabular}[c]{@{}l@{}}Multi-view clustering\end{tabular} & \begin{tabular}[c]{@{}l@{}}BDGP,\\ MNIST,\\ Animal,\\ Caltech101-20,\\ Coil20s\end{tabular}                     & 2022 & ArXive    \\ \hline
	\end{tabular}
	   \begin{tablenotes} 
		\item * denote the method mentioned in the paper is not named. 
	   \end{tablenotes} 
        \end{threeparttable} 
\end{table*}

Wang et al.~\cite{wang2022schemaguidedeventgraphcompletion} proposed a schema-based event graph completion method that simultaneously considers neighborhood and path information when constructing the event schema graph, fully capturing its higher-order graph structure and semantic information. It mitigates the limitations of existing imputation methods and allows the model to forecast missing events.
Zhang et al.~\cite{zhang2022completingnetworkslearninglocal} introduced a model called C-GIN, which infers both missing links and nodes. C-GIN is based on the GAE framework and incorporates a graph isomorphism network, which captures local structural patterns from the observed part of a network and generalizes them to complete the entire graph.
To address the task of partial subgraph learning, where only a partial subgraph is available, Kim et al.~\cite{kim2022modelsbenchmarksrepresentationlearning} introduced the Partial Subgraph Infomax (PSI) framework. This framework maximizes the mutual information between the subgraph summary and node representations of substructures.
For multiplex data, Wang et al.~\cite{wang2022deeppartialmultiplexnetwork} proposed a novel Deep Partial Multiplex Network Embedding (DPMNE) method to handle incomplete multiplex data. It minimizes deep reconstruction loss using GAE, ensures consistency across different views by learning a common latent subspace, and preserves data proximity within the same view using the graph Laplacian.
For incomplete multi-view data, Zhu et al.~\cite{9721669} proposed a novel model for heterogeneous graphs, which flexibly leverages as many incomplete views as possible. The model learns a unified latent representation, effectively balancing consistency and complementarity across different views. To explore the complex relationships between nodes and their representations, the model also introduces neighborhood constraints and view existence constraints to construct heterogeneous graphs.

Moreover, Zhang et al.~\cite{zhang2023learnabletopologicalfeaturesphylogenetic} proposed a topological representation model for phylogenetic inference using the features of graph structure. This method integrates node attributes minimizing Dirichlet energy with GNNs, allowing these features to provide efficient structural information for phylogenetic trees. These features can automatically adapt to many downstream tasks without requiring domain expertise.
Li et al.~\cite{10210121} proposed the Contrastive Sampling-Aggregating Transformer (CSAT) model, which enhances community detection in attribute-missing graphs by integrating graph structures and available node attributes. Leveraging the strengths of transformer models in representation learning, CSAT incorporates contrastive learning to obatin complex relationships between graph structures and node attributes.
To tackle partial multi-view clustering, Wang et al.~\cite{wang2022activeaugmentationfreegraphcontrastivelearning} proposed ACTIVE, a cluster-level contrastive learning framework. ACTIVE constructs relation graphs from existing nodes and propagates inter-instance relationships to missing views, thereby constructing graphs for the missing data. By extending contrastive learning and missing data inference from the instance to the cluster level, ACTIVE mitigates the effect of missing data on clustering.

In general, current label prediction methods for attribute-missing graphs primarily focus on various application tasks. Despite the different characteristics of these applications, a common feature of these methods is the widespread use of GNNs as the core tool. The primary reason for the pivotal role of GNNs in this domain is their capacity to model intricate relationships between nodes in graphs and to mitigate the impact of missing node attributes through information aggregation. As more domains focus on the issue of attribute-missing graphs, GNN-based label prediction approaches are poised for broader development prospects.

\subsection{Summary and discussion}\label{section4.3}	
Attribute-missing graph learning represents an emerging research direction that has garnered significant attention from researchers in recent years. The attribute-missing graphs have recently attracted considerable interest from diverse research communities, leading to numerous publications across various fields. Although there has been progress in developing methods to handle attribute-missing, many challenges remain. These challenges include the development of more robust and scalable algorithms, the ability to handle various types of missing data (e.g., missing nodes, edges, or attributes), and the integration of multiple sources of incomplete information. Additionally, the effectiveness of existing approaches is often domain-dependent, and generalizable frameworks that perform well across different applications are still lacking. Currently, the exploration of attribute-missing problems is in its early stages, making it a promising area deserving further investigation.	

\section{Hybrid-absent graph learning methods }\label{section5}
In the complex graphical structures of the real world, the issue of hybrid-absent graph is prominent due to inherent data incompleteness. Using social networks as an example, a substantial portion of the user population selectively discloses personal information or refrains from sharing, driven by privacy concerns or other factors. This ubiquitous phenomenon of missing information not only undermines the integrity and reliability of the data but also adds significant complexity to tasks such as in-depth analysis, precise prediction, and personalized recommendation, presenting significant challenges to related efforts. In recent years, researchers have introduced some methods for dealing with hybrid-absent graphs. Table \ref{table5} provides a summary of methods for hybrid-absent graphs, and each method is briefly described below.

\begin{table*}[t]
\renewcommand{\arraystretch}{1.2}
	\setlength\tabcolsep{1pt}
	\caption{A summary of methods for hybrid-absent graphs.}
    \label{table5}
	\scriptsize
	\begin{tabular}{p{3.9cm}llp{3.3cm}p{1.3cm}p{1.5cm}}
		\hline
		Model   &  Key Component  &  Applications  &  Datasets                                                                                                                                                                           & Year & Venue                                                                  \\ \hline
		GCNmf~\cite{DBLP:journals/fgcs/TaguchiLM21}   & \begin{tabular}[c]{@{}l@{}}Gaussian mixture model,\\ GCN\end{tabular}          & \begin{tabular}[c]{@{}l@{}}Node classification,\\ Link prediction\end{tabular}               & \begin{tabular}[c]{@{}l@{}}Citation graphs\\ (Cora, Citeseer),\\ Co-purchase graphs\\ (AmaPhoto, AmaComp)\end{tabular}                                                           & 2021 & \begin{tabular}[c]{@{}l@{}}Future\\ Generation \\ Computer \\Systems\end{tabular} \\ \hline
	\multirow{4}{*}{WGNN}\\\cite{chen2022wassersteingraphneuralnetworks}   & \begin{tabular}[c]{@{}l@{}}GNN,\\ Wasserstein aggregation\end{tabular}         & \begin{tabular}[c]{@{}l@{}}Multi-graph matrix\\ completion,\\ Node classification\end{tabular} & \begin{tabular}[c]{@{}l@{}}Citation graphs\\ (Cora, Citeseer, Pubmed),\\ Recommendation systems\\ (Flixster, MovieLens)\end{tabular}                                                & 2021 & ArXive                                                                 \\ \hline
		RITR~\cite{tu2023revisitinginitializingrefiningincomplete}   & \begin{tabular}[c]{@{}l@{}}GNN,\\ Data imputation\end{tabular}                 & \begin{tabular}[c]{@{}l@{}}Attribute completion,\\ Node classification\end{tabular}          & \begin{tabular}[c]{@{}l@{}}Citation graphs\\ (Cora, Citeseer),\\ Co-purchase graphs\\ (AmaPhoto, AmaComp)\end{tabular}                                                           & 2024 &  \begin{tabular}[c]{@{}l@{}}  IEEE \\Transactions\\ on Neural \\Networks\\ and Learning\\ Systems \end{tabular}                                                                 \\ \hline
		\multirow{6}{*}{PCFI}\\\cite{um2023confidencebasedfeatureimputationgraphs}   & \begin{tabular}[c]{@{}l@{}}Graph diffusion,\\ Feature imputation\end{tabular}  & \begin{tabular}[c]{@{}l@{}}Node classification,\\ Link prediction\end{tabular}               & \begin{tabular}[c]{@{}l@{}}Citation networks\\ (Cora, CiteSeer, PubMed), \\ OGBN-Arxiv, \\ Recommendation networks\\ (Amazon-Computers, \\ Amazon-Photo)\end{tabular}                & 2023 & ICLR                                                                   \\ \hline
		\multirow{4}{*}{ASD-VAE}\\\cite{Jiang2024IncompleteGL} & \begin{tabular}[c]{@{}l@{}}GCN,\\ Decoupled variational\\ inference\end{tabular} & \begin{tabular}[c]{@{}l@{}}Attribute completion,\\ Node classification\end{tabular}          & \begin{tabular}[c]{@{}l@{}}Citation graphs\\ (Cora, Citeseer),\\ Co-purchase graphs\\ (AmaPhoto, AmaComp)\end{tabular}                                                           & 2024 & WSDM                                                                   \\ \hline
		PaGNNs~\cite{jiang2021incompletegraphrepresentationlearning}   & \begin{tabular}[c]{@{}l@{}}Partial message \\propagation,\\ GCN\end{tabular}     & Node classification                                                                         & \begin{tabular}[c]{@{}l@{}}Citation graphs\\ (Cora, Citeseer, Cora-ML),\\ Co-purchase graphs\\ (AmaPhoto, AmaComp),\\ Ogbn-arxive,\\ Image dataset\\ (Coil20, FLickr)\end{tabular} & 2021 & \begin{tabular}[c]{@{}l@{}}ArXive\end{tabular}  \\ \hline
	\end{tabular}
\end{table*}

Taguchi et al.~\cite{DBLP:journals/fgcs/TaguchiLM21} first proposed a Graph Convolutional Network (GCNmf) model that simultaneously addresses the issues of both attribute-missing and attribute-incomplete graphs. The model integrates the handling of missing attributes with graph learning techniques within its neural network framework. Specifically, it encodes absent attributes and estimates the expected activations for the first-layer neurons in the GCN model. Experimental data demonstrate that the GCNmf model exhibits robust stability in dealing with incomplete graphs while outperforming traditional imputation-based methods in tasks such as node classification and link prediction.
Jiang et al.~\cite{jiang2021incompletegraphrepresentationlearning} proposed a variant of the GCN called PaGCN, which draws on the message-passing scheme of traditional GCNs but introduces adaptive adjustments to maintain simplicity and efficiency, similar to standard GCNs. Notably, PaGCN seamlessly handles graph data with missing attributes, eliminating the need for imputing or predicting missing values, thereby streamlining the workflow. Furthermore, using PaGCN, the authors introduced a novel dropout technique aimed at optimizing the training process of GCNs.
Chen et al.~\cite{chen2022wassersteingraphneuralnetworks} introduced the Wasserstein GNN (WGNN) framework, which aims to fully utilize the limited available attribute observations and obtain the uncertainty caused by missing attributes. WGNN maps the nodes to low-dimensional probability distributions based on the decomposition of the attribute matrix. Additionally, the network introduces a message-passing mechanism that aggregates distributional information from neighboring nodes in the Wasserstein metric space, greatly improving the model's ability to obtain complex graph structures.
In view of the significant decline in the performance of existing methods for handling incomplete graphs under conditions of a high missing feature rate, Um et al.~\cite{um2023confidencebasedfeatureimputationgraphs} proposed a Pseudo-Confidence-based Feature Imputation (PCFI) method that can address both attribute-missing and attribute-incomplete graphs. PCFI introduces channel-wise confidence in node features, assigning a certainty metric to each imputed feature. By incorporating this confidence into the imputation process, the approach provides a more accurate assessment of imputation reliability, aiding subsequent tasks. Additionally, PCFI uses the channel-wise shortest path distance between nodes with missing features and their nearest nodes with known features to create a pseudo-confidence measure. This strategy remains stable even with missing rates as high as 99.5\%.
Jiang et al.~\cite{Jiang2024IncompleteGL} proposed the ASD-VAE model, which draws inspiration from brain cognitive processes and multimodal fusion, parameterizing the shared latent space through a coupled-decoupled learning process. ASD-VAE independently encodes graph attributes and structure, creating separate feature representations for each view. It then learns a shared latent space by maximizing the likelihood of their joint distribution through coupling. A decoupling operation separates the shared space into individual views, with reconstruction loss computed for each view, and missing attribute values imputed from the shared latent space.
Moreover, Tu et al.~\cite{tu2023revisitinginitializingrefiningincomplete} leveraged the ITR algorithm framework and proposed the RITR model, which adopts a sequential imputation strategy: first initializing to build the foundation, then refining to enhance quality. This approach effectively achieves accurate imputation for attribute-incomplete and attribute-missing nodes, offering new solutions to incomplete graph data analysis.

\subsection{Summary and discussion}\label{section5.1}
Compared to the problems of attribute-missing and attribute-incomplete graphs, hybrid-absent graphs present a more challenging problem. Although this issue has attracted some attention from researchers, relevant studies remain limited.  Given the widespread occurrence of hybrid-absent graphs in real-world scenarios, this paper aims to raise awareness among researchers and promote further investigation into this problem, with the goal of developing more effective methods for addressing it.

\begin{table*}[]
\renewcommand{\arraystretch}{1.3}
\caption{Summary of non-graph structured datasets.}
\label{table6}
\scriptsize

\begin{tabular}{p{1.3cm}p{2.3cm}p{2.0cm}p{1.5cm}l}
\hline
\multicolumn{1}{l}{Type} & \multicolumn{1}{l}{Task}  & \multicolumn{1}{l}{Dataset} & \multicolumn{1}{l}{Source} & \multicolumn{1}{l}{Paper} \\ \hline
\multirow{30}{*}{UCI}     & Feature estimation, Label prediction & concrete                    &                         & IGRM~\cite{DBLP:conf/aaai/ZhongGY23}, GRAPE 
               \\ 
& Feature estimation, Label prediction         &     housing      &                            & IGRM, GRAPE               \\
& Feature estimation, Label prediction & wine                        &                            & IGRM, GRAPE               \\
&Feature estimation, Label prediction & Yacht                       &                            & IGRM, GRAPE               \\
&Feature estimation, Label prediction & protein                     &                            & GRAPE~\cite{2020Handling}           \\
& Feature estimation & heart                       &                            & IGRM                      \\
&Feature estimation & DOW30                       &                            & IGRM                      \\
&Feature estimation & E-commerce                  &                            & IGRM                      \\
&Feature estimation & Diabetes                    &                            & IGRM                      \\
&Feature estimation & energy                      &                            & GRAPE                     \\
& Feature estimation & kin8nm                      &                            & GRAPE                     \\
& Feature estimation & naval                       &                            & GRAPE                     \\
& Feature estimation & power                       &                            & GRAPE                     \\
&Feature estimation & abalone                     &                            & GINN~\cite{DBLP:journals/nn/SpinelliSU20}          \\
&Feature estimation & anuran-calls                     &                            & GINN            \\
&Feature estimation & balance-scale                     &    \cite{2007UCI}                         & GINN           \\
&Feature estimation & breast-cancer-diagnostic                      &                            & GINN           \\
&Feature estimation & car-evaluation                      &                            & GINN           \\
&Feature estimation & default-credit-card                     &                            & GINN           \\
&Feature estimation & electrical-grid-stability                     &                            & GINN           \\
&Feature estimation & ionosphere                     &                            & GINN           \\
&Feature estimation & iris                     &                            &GINN           \\
&Feature estimation & page-blocks                     &                            & GINN          \\
&Feature estimation & phishing                     &                            & GINN          \\
&Feature estimation & satellite                      &                            & GINN          \\
&Feature estimation & tic-tac-toe                      &                            & GINN          \\
&Feature estimation & turkiye-student-evaluation                      &                            & GINN          \\
&Feature estimation & wireless-localization                      &                            & GINN          \\
&Feature estimation & yeast                      &                            & GINN          \\
\hline
\begin{tabular}{@{} l @{}} 
        Recommen-\\dation
\end{tabular}                                       
    &Feature estimation, Node & MovieLens                   & \cite{DBLP:journals/corr/KingmaB14} & \begin{tabular}[c]{@{}l@{}}MC~\cite{DBLP:journals/focm/CandesR09}, \\GMC\\\cite{kalofolias2014matrix},\\ RGCNN~\cite{DBLP:conf/nips/MontiBB17}, \\CGMC~\cite{2018Convolutional}\\ IGMC~\cite{zhang2020inductivematrixcompletionbased}, \\IMC-GAE~\cite{DBLP:conf/cikm/0005ZTZHD021},\\WGNN~\cite{chen2022wassersteingraphneuralnetworks}, \\GC-MC~\cite{2017Graph}\end{tabular}  \\
\hline
\end{tabular}
\end{table*}

\section{Incomplete graph learning in practice}\label{section6}
In this section, we provide a detailed list of datasets related to incomplete graph learning methods. The rationales are delineated below: (1) to validate the performance of these methods; (2) to further investigate the practical utility and applicability of incomplete graph learning and promote the development of benchmark datasets for this field (currently, incomplete graph learning primarily relies on existing complete graphs, and no benchmark datasets have been established yet). Additionally, we summarize incomplete processing mode as well as evaluation tasks and applications frequently used in incomplete graph learning methods.

\subsection{Datasets}\label{section6.1}
In reviewing the incomplete graph learning methods, we observe that current research primarily relies on two categories of datasets: non-graph structured datasets and graph structured datasets. Non-graph structured datasets refer to those in which the nodes are independent and do not contain any edges between them. The details of these two types of datasets are as follows. 

\subsubsection{Non-graph structured datasets}\label{section6.1.1}
\begin{figure*}[ht]
\centering
\includegraphics[width=0.9\columnwidth]{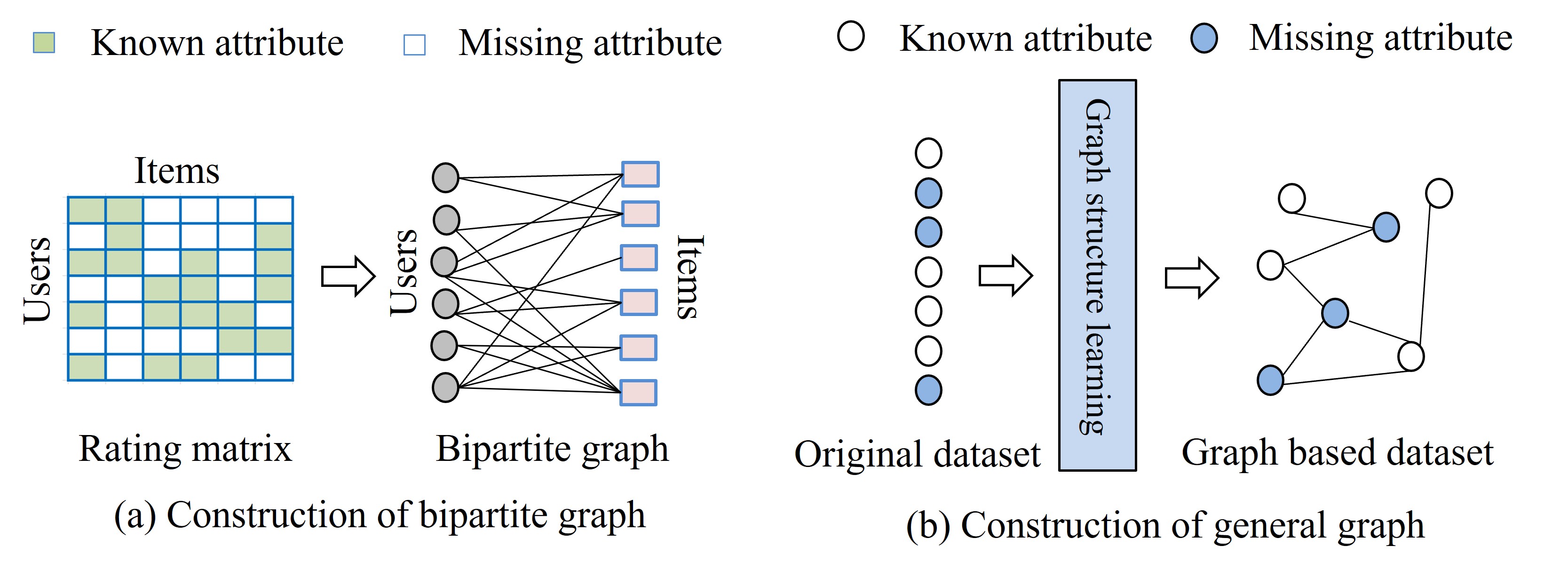}
\caption{Construction of graphs on non-graph structured datasets. }
\label{Fig13}
\end{figure*}

Most non-graph structured datasets are utilized to evaluate attribute-incomplete graph learning algorithms, particularly those that leverage data imputation strategies. These datasets typically transform discrete data points into graph-structured data by constructing virtual edges, as shown in Figure \ref{Fig13}. This transformation helps to complete and recover the missing attributes in the original graph. Table \ref{table6} provides a summary of non-graph structured datasets used in incomplete graph learning.
The non-graph structured datasets used for incomplete graph learning primarily focus on the UCI machine learning repository~\cite{2007UCI} as well as recommendation systems. The UCI datasets span various domains, including computer science, civil engineering, biology, medicine, and thermodynamics. Each dataset consists of numerous nodes and features. For example, the YACHT dataset contains 314 nodes and 6 features, while the PROTEIN dataset contains more than 45,000 observations and 9 features. Detailed information on UCI datasets used in incomplete graph learning is available in~\cite{2020Handling, DBLP:conf/aaai/ZhongGY23, DBLP:journals/nn/SpinelliSU20}. For recommendation systems, researchers utilize the MovieLens \footnote{https://grouplens.org/datasets/movielens/} dataset as a non-graph structured dataset to validate incomplete graph learning. The MovieLens dataset serves as a valuable resource for research in recommendation systems. It contains a large volume of movie rating data, making it a crucial resource for studying personalized recommendation algorithms. MovieLens datasets used in incomplete graph learning are detailed in~\cite{2017Graph, zhang2020inductivematrixcompletionbased, 2020Handling, DBLP:conf/nips/MontiBB17, 2018Convolutional, zhang2020inductivematrixcompletionbased, 2014Multiple, DBLP:conf/cikm/0005ZTZHD021,kalofolias2014matrix}.

\subsubsection{Graph structured datasets}\label{section6.1.2}
Graph-structured datasets are widely applied in incomplete graph learning. Researchers have used various types of graph-structured data to validate the effectiveness of proposed methods. A summary of graph-structured datasets is provided in Table \ref{table7}. These datasets span a wide range of application domains, including citation networks, social networks, purchase networks, recommendation systems, biochemical networks, and others. This subsection details the most commonly used datasets.

\textbf{Citation networks.}~In citation networks, nodes denote individual articles, while the edges denote the citation relationships between them. Cora~\cite{2012Query}, Pubmed~\cite{Prithviraj2008Collective}, and Citeseer~\cite{DBLP:conf/icml/LuG03} are the most commonly used citation networks for validating incomplete graph learning methods. Among these datasets, the Pubmed contains 44,338 citation links and 19,717 diabetes-related publications. These publications are classified into 3 distinct categories, with each publication consisting of a set of 500 topics. The Cora dataset, contains 5,429 citation links and 2,708 scientific publications, covering 7 different categories. Each publication in this dataset is represented by 1,433 topics. Finally, the Citeseer dataset contains 4,732 citation links and 3,327 scientific publications, systematically classified into six distinct categories. Each publication within this dataset is characterized by 3,703 distinct topics.

\textbf{Co-occurrence networks.}~In co-purchase networks, the nodes denote commodities and the edges represent that these commodities are purchased frequently together. In coauthor networks, nodes represent authors, and edges indicate that two authors are linked if they have jointly authored a publication. The AmaPhoto, AmaComp \footnote{https://docs.dgl.ai/api/python/dgl.data.html\#amazon-co-purchase-dataset}, and Coauthor-CS~\cite{DBLP:journals/corr/abs-1811-05868} datasets are the most commonly used co-occurrence networks for validating incomplete graph learning methods. Among these datasets, the AmaComp dataset consists of 245,861 links and 13,752 commodities, which belong to 10 categories. Each commodity consists of 767 product evaluations. Meanwhile, the AmaPhoto dataset consists of 119,081 links and 7,650 commodities, which are categorized into 8 groups. Each commodity in this dataset consists of 745 product evaluations. Additionally, the Coauthor-CS dataset encompasses 18,333 nodes interconnected by 81,894 edges. The node attributes represent the frequencies of 6,805 keywords in the articles of each author.

\afterpage{\clearpage}
\renewcommand{\arraystretch}{1.2}
{\footnotesize
\begin{longtable}{p{2cm}p{2.5cm}p{2.0cm}p{3.3cm}p{4.5cm}}
\captionsetup{justification=raggedright, singlelinecheck=false}
\caption{Summary of graph structured datasets.} \label{table7} \\

\hline
\raggedright Type & Task & Dataset & Source & Paper \\ \hline
\endfirsthead

\hline
\raggedright Type & Task & Dataset & Source & Paper \\ \hline
\endhead
\hline
		\multirow{15}{*}{\begin{tabular}[c]{@{}l@{}}Citation\\ networks\end{tabular}}       & \begin{tabular}[c]{@{}l@{}}Node/Link,\\ Feature estimation\end{tabular}  & Cora                        & \cite{Prithviraj2008Collective}                   & \begin{tabular}[c]{@{}l@{}} MEGAE~\cite{DBLP:conf/aaai/GaoNCTLXZTL23}, \\GCNmf~\cite{DBLP:journals/fgcs/TaguchiLM21},
        \\T2-GNN~\cite{DBLP:conf/aaai/Huo0LHYW23}, \\PaGCN~\cite{10495099}\\
		PA-GNN \\\cite{tang2020transferring}, \\FP~\cite{DBLP:conf/log/RossiK0C0B22}, \\SGHFP~\cite{DBLP:conf/icassp/LeiFWQHPY23}, \\SAT~\cite{Chen_2022}, \\WGNN~\cite{chen2022wassersteingraphneuralnetworks}\\Amer~\cite{9765782}, \\ITR~\cite{DBLP:conf/ijcai/TuZLLCZZC22}, \\SVGA~\cite{yoo2023accuratenodefeatureestimation}, \\AmGCL~\cite{zhang2023amgclfeatureimputationattribute}, \\RITR~\cite{tu2023revisitinginitializingrefiningincomplete}\\ AIAE~\cite{XIA2024111583}, \\CSAT~\cite{10210121}, \\C-GIN~\cite{zhang2022completingnetworkslearninglocal},\\ DPMNE~\cite{wang2022deeppartialmultiplexnetwork}\\ PCFI~\cite{um2023confidencebasedfeatureimputationgraphs}, \\ASD-VAE~\cite{Jiang2024IncompleteGL},\\ IANRW~\cite{DBLP:journals/ida/WeiPHHYLZ19} \end{tabular} \\ \cline{2-5} 
		&\begin{tabular}[c]{@{}l@{}}\multirow{1.5}{*}{Node/Link}\end{tabular}  & \multirow{1.5}{*}{Cora-ML} & \multicolumn{1}{l}{}     & \multirow{1.5}{*}{PaGCN}   \vspace{2pt}  \\ \cline{2-5} 
		& \begin{tabular}[c]{@{}l@{}}Node/Link,\\Feature estimation \end{tabular}   & Citeseer                    & \cite{Prithviraj2008Collective} & \begin{tabular}[c]{@{}l@{}} MEGAE, GCNmf, T2-GNN, IANRW\\
			PA-GNN, FP, SGHFP, SAT\\ Amer, ITR, SVGA, AmGCL\\AIAE, CSAT, RITR, PCFI, ASD-VAE \end{tabular} \\ \cline{2-5} 
		& \begin{tabular}[c]{@{}l@{}}Node,\\ Feature estimation\end{tabular} & Pubmed                      & \cite{Prithviraj2008Collective} &\begin{tabular}[c]{@{}l@{}} MEGAE, T2-GNN, PA-GNN, FP,\\ SGHFP, SAT, Amer, SVGA, AmGCL,\\ CSAT, WGNN, PCFI\end{tabular} \\ \cline{2-5} 
		&\multirow{2}{*}{Node/Link} &\multirow{2}{*} {DBLP}    & \cite{2020One2Multi} & \begin{tabular}[c]{@{}l@{}} \multirow{1}{*}{MCGC~\cite{DBLP:conf/nips/PanK21},} \\\multirow{1}{*}{HGCA~\cite{9724614},} \\\multirow{1}{*}{HGNN-AC~\cite{10.1145/3442381.3449914},}\\\multirow{1}{*}{AutoAC~\cite{zhu2023autoacautomatedattributecompletion},}\\ \multirow{1}{*} { HetReGAT-FC~\cite{LI2023424},}\\\multirow{1}{*} {DPMNE~\cite{wang2022deeppartialmultiplexnetwork}}\end{tabular}\\ \cline{2-5} 
		& \multirow{1.5}{*}{Node/Link} & \multirow{1.5}{*}{OGBN-ArXiv}                  &  \multirow{1.5}{*}{\cite{DBLP:journals/corr/abs-1811-05868}} & \multirow{1.5}{*}{FP, PaGCN, PCFI } \vspace{4pt}\\ \cline{2-5} 
		& \multirow{1.5}{*}{Node/Link} & \multirow{1.5}{*}{ArXive}                      & \multirow{1.5}{*}{\cite{yoo2023accuratenodefeatureestimation}} & \multirow{1.5}{*}{SVGA} \vspace{4pt} \\ \hline
        \multirow{6}{*}{\begin{tabular}[c]{@{}l@{}}Co-purchase \\ graphs\end{tabular}}     & \begin{tabular}[c]{@{}l@{}} \multirow{2}{*}{Node/Link}\end{tabular}  & \multirow{2}{*}{Amac}                        & \cite{10.1162/qss_a_00021} &{\begin{tabular}[c]{@{}l@{}} GCNmf, PA-GNN, FP, MCGC, ITR,\\ AIAE, CSAT, PaGCN, RITR, PCFI,\\ \end{tabular}  \begin{tabular}[c]{@{}l@{}} \raggedright ASD-VAE \end{tabular}}   \\ \cline{2-5} 
		& \begin{tabular}[c]{@{}l@{}}Node/Link,\\ Feature estimation\end{tabular}  & Amap                        & \cite{10.1162/qss_a_00021} & \begin{tabular}[c]{@{}l@{}}GCNmf, PA-GNN, FP, SGHFP\\ ASD-VAE, MCGC, SAT, Amer, ITR\\ SVGA, PCFI, AmGCL, AIAE\\ CSAT, PaGCN, RITR \end{tabular} \\ \hline
        \begin{tabular}[c]{@{}l@{}} \multirow{1}{*}{Co-occurence}\\ \multirow{1}{*}{network} \end{tabular} &\multirow{1}{*}{Node} & \multirow{1}{*}{Actor}
		& \multirow{1}{*}{\cite{DBLP:conf/ijcai/OrsiniFR15}} &\multirow{1}{*}{PA-GNN} 
		\\ \cline{2-5}  \hline
		\begin{tabular}[c]{@{}l@{}}\multirow{3}{*}{Wikipedia}\\ \multirow{3}{*}{networks} \end{tabular}    & \begin{tabular}[c]{@{}l@{}} \multirow{2}{*}{Node}\end{tabular}  & \multirow{2}{*}{Chameleon}                        & \cite{9514682} & \multirow{2}{*}{T2-GNN, PA-GNN} \\ \cline{2-5} 
		& \begin{tabular}[c]{@{}l@{}} \multirow{1.5}{*}{Node}\end{tabular}  &  \multirow{1.5}{*}{Squirrel}                        &  \multirow{1.5}{*}{\cite{9514682}} &  \multirow{1.5}{*}{T2-GNN, PA-GNN} \vspace{4pt}\\ \hline
        \multirow{5}{*}{\begin{tabular}[c]{@{}l@{}}Webpage\\ networks \end{tabular}}     & \begin{tabular}[c]{@{}l@{}} \multirow{3}{*}{Node} \end{tabular}  & \multirow{3}{*}{Cornell}                        & \cite{DBLP:journals/tfs/Garcia-PlazaFMZ17} & \multirow{3}{*}{T2-GNN} \\ \cline{2-5} 
		& \begin{tabular}[c]{@{}l@{}}  \multirow{1.5}{*}{Node} \end{tabular}  &  \multirow{1.5}{*}{Wisconsin}                        &  \multirow{1.5}{*}{\cite{DBLP:journals/tfs/Garcia-PlazaFMZ17}} &  \multirow{1.5}{*}{T2-GNN} \vspace{4pt}\\ \cline{2-5} 
		& \begin{tabular}[c]{@{}l@{}} \multirow{1.5}{*}{Node} \end{tabular}  &  \multirow{1.5}{*}{Texas}                        &  \multirow{1.5}{*}{\cite{DBLP:journals/tfs/Garcia-PlazaFMZ17}} &  \multirow{1.5}{*}{T2-GNN} \vspace{4pt}\\ \hline
        \multirow{11}{*}{\begin{tabular}[c]{@{}l@{}}Recommendation \\ systems\end{tabular}}     & \begin{tabular}[c]{@{}l@{}} Node,\\Feature estimation\end{tabular}  & Flixster                        & \cite{DBLP:conf/recsys/JamaliE10} &{\begin{tabular}[c]{@{}l@{}} GMC~\cite{kalofolias2014matrix}, \\RGCNN~\cite{DBLP:conf/nips/MontiBB17}, \\CGMC~\cite{2018Convolutional}, \\IANRW,IGMC~\cite{zhang2020inductivematrixcompletionbased}, \\IMC-GAE~\cite{DBLP:conf/cikm/0005ZTZHD021}, \\Feras~\cite{DBLP:journals/corr/abs-2210-01803},\\ CSAT, DPMNE, PaGCN, WGNN,\\ MC~\cite{DBLP:journals/focm/CandesR09} \end{tabular}}    \\ \cline{2-5}
		& \multirow{2}{*}{Feature estimation} &\multirow{2}{*} {Douban}                    & \cite{DBLP:conf/wsdm/MaZLLK11} & \begin{tabular}[c]{@{}l@{}}\multirow{2}{*}{GMC, RGCNN, CGMC}\\ \multirow{2}{*}{IGMC, IMC-GAE, MC}\end{tabular}  \\ \cline{2-5} 
		&\multirow{2}{*}{Feature estimation} &\multirow{2}{*}{YahooMusic}                    & \cite{DBLP:journals/jmlr/DrorKKW12} & \begin{tabular}[c]{@{}l@{}}\multirow{2}{*}{MC, GMC, RGCNN, CGMC}\\ \multirow{2}{*}{IGMC, IMC-GAE}\end{tabular} \\ \cline{2-5} 
		& Node/Link & Poliblog                    & \cite{10.1145/1134271.1134277} & \begin{tabular}[c]{@{}l@{}}Matrix Completion with \\Hierarchical
		Graph Side Information\end{tabular} \\ \cline{2-5}
		&  \multirow{1.5}{*}{Node/Link} &  \multirow{1.5}{*}{Yelp}                        &  \multirow{1.5}{*}{\cite{DBLP:conf/aaai/LuSH019}} &  \multirow{1.5}{*}{HGCA, CSAT} \vspace{4pt}\\ \cline{2-5}
		&  \multirow{1.5}{*}{Node/Link} &  \multirow{1.5}{*}{LastFM}                      &  \multirow{1.5}{*}{\cite{zhu2023autoacautomatedattributecompletion}} &  \multirow{1.5}{*}{AutoAC, DPMNE} \vspace{4pt}\\ \cline{2-5}
	  \hline
		\multirow{7}{*}{\begin{tabular}[c]{@{}l@{}}Bioinformatics\end{tabular}}     & \begin{tabular}[c]{@{}l@{}}\multirow{3}{*}{Graph,}\\\multirow{3}{*}{Feature estimation}\end{tabular}  &\multirow{3}{*}{PROTEINS\_full}                        & \cite{DBLP:conf/ismb/BorgwardtOSVSK05} &{\begin{tabular}[c]{@{}l@{}} \multirow{3}{*}{MEGAE}\end{tabular}}    \\ \cline{2-5}
		&  \multirow{1.5}{*}{RMSE, Graph/Node} &  \multirow{1.5}{*}{PPI}                    &  \multirow{1.5}{*}{\cite{DBLP:conf/ismb/BorgwardtOSVSK05}} &  \multirow{1.5}{*}{Feras} \vspace{4pt}\\ \cline{2-5} 
		& \begin{tabular}[c]{@{}l@{}}\multirow{3}{*}{Graph,}\\ \multirow{3}{*}{Feature estimation}\end{tabular}  & \multirow{3}{*}{ENZYMES}    & \cite{DBLP:journals/nar/SchomburgCEGHHS04} & \begin{tabular}[c]{@{}l@{}}\multirow{3}{*}{MEGAE} \end{tabular} \\ \hline
		\multirow{5}{*}{\begin{tabular}[c]{@{}l@{}}Chemistry \end{tabular}} & \multirow{2}{*}{Feature estimation} & \multirow{2}{*}{QM9}                    & \cite{2014Quantum} & \multirow{2}{*}{MEGAE} \\ \cline{2-5} 
		& \multirow{2}{*}{Feature estimation} & \multirow{2}{*}{FIRSTMM\_DB}                   & \cite{2013Graph} &\multirow{2}{*}{MEGAE} \\ \hline
		\multirow{1.5}{*}{\begin{tabular}[c]{@{}l@{}}Computer vision \end{tabular}} & \multirow{1.5}{*}{Feature estimation} &  \multirow{1.5}{*}{FRANKENSTEIN}
		                    &  \multirow{1.5}{*}{\cite{DBLP:conf/ijcai/OrsiniFR15}} &  \multirow{1.5}{*}{MEGAE} \vspace{4pt} 
        \\ \hline
        \multirow{4}{*}{\begin{tabular}[c]{@{}l@{}}Twitch gamers \end{tabular}} &\multirow{1.5}{*}{Node} & \multirow{1.5}{*}{Twitch}
        & \multirow{1.5}{*}{\cite{9514682}} & \multirow{1.5}{*}{PA-GNN} \vspace{4pt} 
        \\ \cline{2-5}
        &\multirow{3}{*}{Node} & \multirow{3}{*}{OGBN-Products}
        & \cite{2020Open} &\multirow{3}{*}{FP}
        \\ \hline
		\multirow{9}{*}{\begin{tabular}[c]{@{}l@{}}Others\end{tabular}}     & \begin{tabular}[c]{@{}l@{}} Node/Link,\\Feature estimation\end{tabular}  & Steam                        & \cite{Chen_2022} &{\begin{tabular}[c]{@{}l@{}}  SAT, SVGA, AmGCL\end{tabular} }   \\ \cline{2-5}
		    & \begin{tabular}[c]{@{}l@{}} Node/Link,\\Feature estimation\end{tabular}  &  Coauther-CS                         & \cite{DBLP:journals/corr/abs-1811-05868} &{\begin{tabular}[c]{@{}l@{}}  SAT, Amer, SVGA, AmGCL, CSAT,\\ C-GN\end{tabular} }   \\ \cline{2-5} 
		& Node & ACM                         & \cite{2020One2Multi} & \begin{tabular}[c]{@{}l@{}}MCGC, HGCA, HGNN-AC, AutoAC, \\HetReGAT-FC\end{tabular} \\ \cline{2-5} 
		& Node & IMDB                        & \cite{2020One2Multi} & \begin{tabular}[c]{@{}l@{}}MCGC, HGNN-AC, AutoAC,\\ HetReGAT-FC \end{tabular}\\ \cline{2-5}  
		& \multirow{1.5}{*}{Node} & \multirow{1.5}{*}{Reddit}                        & \multirow{1.5}{*}{\cite{DBLP:conf/nips/HamiltonYL17}} & \multirow{1.5}{*}{Feras, CSAT} \vspace{4pt} \\ \hline
\end{longtable}
}
\vspace{-10pt}
\noindent In this table, node represents node tasks, such as node classification and node clustering. The link represents link tasks, such as link prediction.

\vspace{0.5cm}
\textbf{Wikipedia networks.}~Chameleon and Squirrel~\cite{9514682} are two specialized topic-based networks within Wikipedia, structured as page-to-page connections. In these networks, each node represents an individual webpage, while the edges depict bidirectional links that exist between those pages. The distinguishing attributes of each node correspond to various informative terms found within the respective Wikipedia pages. Among these datasets, the Chameleon dataset contains 31,421 links and 2,277 nodes, while the Squirrel dataset contains 198,493 links and 5,201 nodes. The labels of these datasets correspond to five categories based on the average monthly traffic of the webpages.

\textbf{Webpage networks.}~The Wisconsin, Cornell, and Texas datasets~\cite{DBLP:journals/tfs/Garcia-PlazaFMZ17} are webpage networks collected by Carnegie Mellon University. In these datasets, the nodes represent web pages, and the edges correspond to hyperlinks between them. The node features are represented by the bag-of-words model of the web pages. Among these, the Wisconsin dataset contains 499 links and 251 nodes, with each node having 1,703 attributes. The Cornell dataset includes 295 links and 183 nodes, and similarly, each node is represented by 1,703 attributes. The Texas dataset has 309 links and 183 nodes, with the same number of attributes per node.

\textbf{Recommendation graphs.}~Flixster, Douban, and YahooMusic \footnote{https://github.com/fmonti/mgcnn} are the most commonly used recommendation graphs for validating incomplete graph learning methods. These datasets contain user and item-side information in the form of graphs. Flixster is a movie-focused social networking site where users share ratings of films, discuss new releases, and connect with like-minded individuals. Douban is a website that allows users to share reviews and opinions about movies. The dataset includes over two million short reviews related to 28 movies. The YahooMusic dataset is widely used in music recommendation research, encompassing user ratings for musical tracks, albums, artists, and genres.

\textbf{Bioinformatics and chemistry graphs.}~For bioinformatics, PROTEINS\_full~\cite{DBLP:conf/ismb/BorgwardtOSVSK05}, PPI~\cite{DBLP:conf/ismb/BorgwardtOSVSK05}, and ENZYMES~\cite{DBLP:journals/nar/SchomburgCEGHHS04} are commonly used to evaluate the effectiveness of existing methods. PROTEINS\_full is a dataset containing proteins, classified as enzymes and non-enzymes based on their functional properties. In this dataset, nodes represent amino acids, and edges connect two nodes if the distance between them is less than 6 Angstroms. The ENZYMES dataset is a collection of graph-based data constructed from biomolecular protein structures, consisting of 600 graphs, each representing one of six distinct protein structures. The PPI dataset contains 24 graphs, each representing a different human tissue. On average, each graph contains 2,371 nodes, for a total of 56,944 nodes and 818,716 edges. Each node is represented by a 50-dimensional feature vector, encompassing positional gene sets, motif sets, and immunological features. Gene Ontology terms are used as labels, with a total of 121 categories. For chemistry, QM9~\cite{2014Quantum} and FIRSTMM\_DB~\cite{2013Graph} are commonly used for evaluating the effectiveness of methods. The QM9 dataset includes the composition, spatial information, and corresponding properties of 130,000 organic molecules.

In addition to the aforementioned datasets, there are other datasets employed for validating and evaluating incomplete graph learning methods. For instance, when dealing with heterogeneous graphs, datasets such as ACM, IMDB~\cite{2020One2Multi}, and Yelp~\cite{DBLP:conf/aaai/LuSH019} are frequently used to verify the effectiveness of the methods. Furthermore, there exist datasets tailored for various application domains, including but not limited to traffic prediction and knowledge graphs. This section focuses on introducing commonly used datasets, providing a foundational understanding for researchers. For those interested in delving deeper into specific application domains, it is advisable to consult the relevant literature, which provides detailed insights and analyses tailored to the unique challenges and opportunities in those fields.

\subsection{Incomplete processing mode}\label{section6.2}
Based on the absence of graph attributes, the existing incomplete graphs can be classified as: (1) attribute-incomplete graphs~\cite{DBLP:journals/fgcs/TaguchiLM21,DBLP:journals/kbs/KongZSZLY23,10495099,Chen_2022,DBLP:journals/corr/abs-2408-04845,guo2023fairattributecompletiongraph,DBLP:journals/corr/abs-2407-18170,Jiang2024IncompleteGL}. These graphs have nodes missing some attributes but retaining partial information (Figure \ref{Fig3}(a)). (2) Attribute-missing graphs~\cite{DBLP:journals/fgcs/TaguchiLM21,DBLP:journals/kbs/KongZSZLY23,10495099,9765782,DBLP:conf/ijcai/TuZLLCZZC22,zhang2023amgclfeatureimputationattribute,XIA2024111583,Jiang2024IncompleteGL}. These graphs have some nodes that completely lack attributes (Figure \ref{Fig3}(b)). (3) Hybrid-missing graphs~\cite{chen2022wassersteingraphneuralnetworks,tu2023revisitinginitializingrefiningincomplete,um2023confidencebasedfeatureimputationgraphs,Jiang2024IncompleteGL,jiang2021incompletegraphrepresentationlearning}. These graphs consist of both attribute-incomplete and attribute-missing nodes, which have complex effects on graph analysis (Figure \ref{Fig3}(c)).
Researchers preprocess datasets by introducing a missing rate to simulate missing attributes. Details of the incomplete data processing are provided below.

\textbf{Uniform missing}.
For each node, a fraction of the attributes is randomly selected and removed from the attribute matrix $\mathbf{X}$ according to a missing rate $\rho \in \left \{  0.1, 0.2, ..., 0.9 \right \} $. This method simulates scenarios where node attributes are partially missing or unavailable (i.e., attribute-incomplete graphs). During the removal process, the attributes are randomly removed according to $\rho = |L| / (ND)$, where $|L|$ denotes the number of attributes to be removed from each node, and $ND$ is the total number of attributes of a node, ensuring uniformity in the selection. 

\begin{figure*}[ht]
\centering
\includegraphics[width=0.7\textwidth]{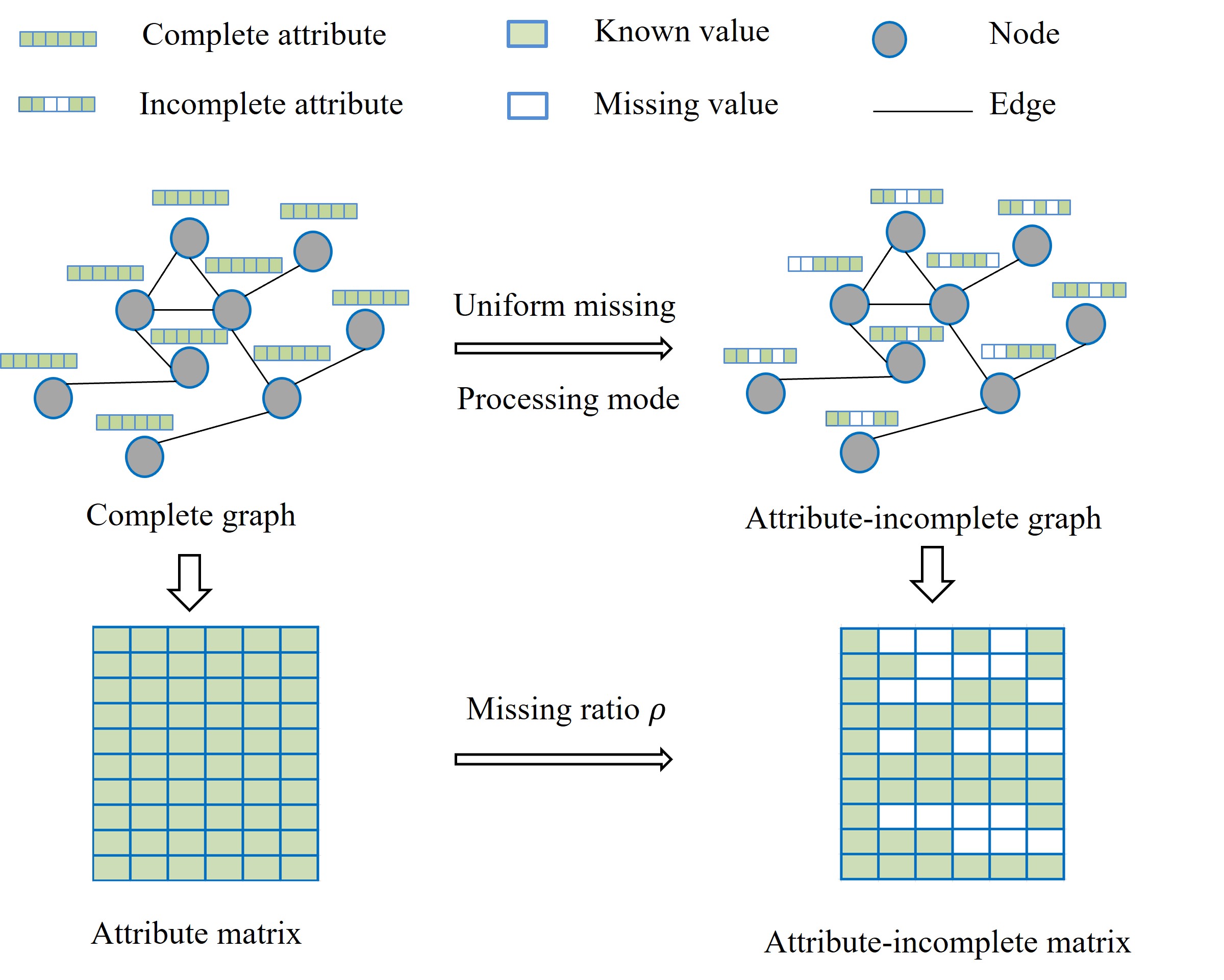}
\caption{The incomplete processing mode of attribute-incomplete graphs. }
\label{Fig14}
\end{figure*}
Based on the uniform missing pattern, researchers randomly select a subset of nodes from the entire node set according to a predefined ratio $r_m$. For the selected nodes, feature values are randomly set to missing according to the missing rate $\rho$, typically represented as zeros, thereby simulating the impact of attribute incompleteness on graph analysis. A visual representation of this process is shown in Figure \ref{Fig14}.

\textbf{Structurally missing}. 
For attribute-missing graphs, attributes are removed from the attribute matrix $\mathbf{X}$ for a subset of nodes determined by the missing rate $\pi$. This simulates a realistic scenario where the attributes of the selected nodes are completely missing. Specifically, a subset ${\mathbf{V} }' \subset \mathbf{V}$ is randomly selected with uniform probability such that $\pi = \left | \mathbf{V}' \right | /N$. $N$ represents the number of nodes in the dataset. 

\begin{figure*}[ht]
\centering
\includegraphics[width=0.7\textwidth]{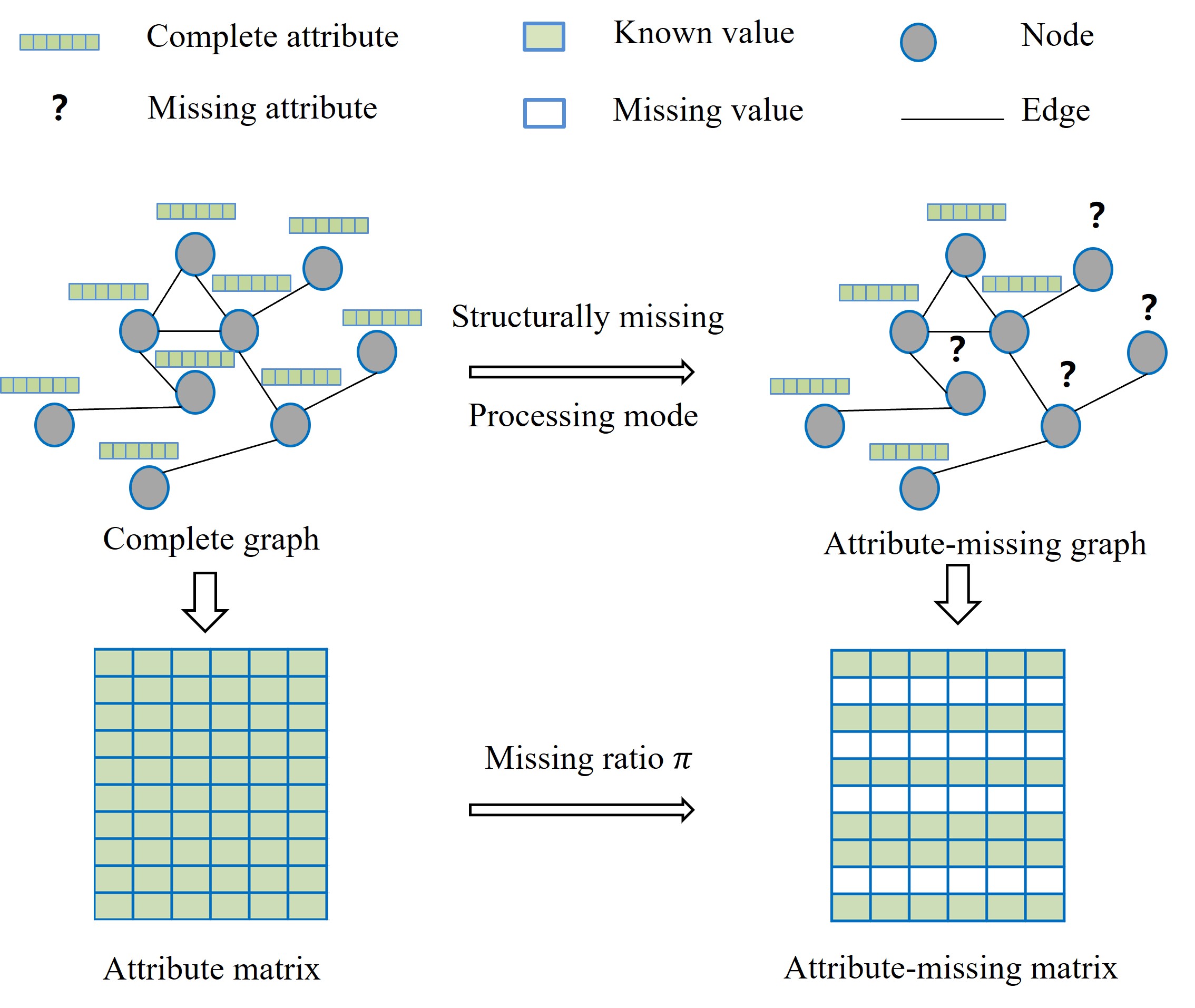}
\caption{ The incomplete processing mode of attribute-missing graphs.}
\label{Fig15}
\end{figure*}
Based on the structurally missing pattern, researchers randomly select a subset of nodes from the entire node set according to a predefined ratio $\pi$, and set the values of the selected attributes to be missing, typically represented by zeros to indicate missing values. A visual representation of this process is shown in Figure \ref{Fig15}.

\textbf{Hybrid missing}.
For the hybrid missing pattern, both uniform missing and structurally missing modes are applied simultaneously to remove attributes from the matrix $\mathbf{X}$. Specifically, attributes are removed based on the missing rates $\rho$ and $\pi$, respectively. 

\begin{figure*}[ht]
\centering
\includegraphics[width=0.7\textwidth]{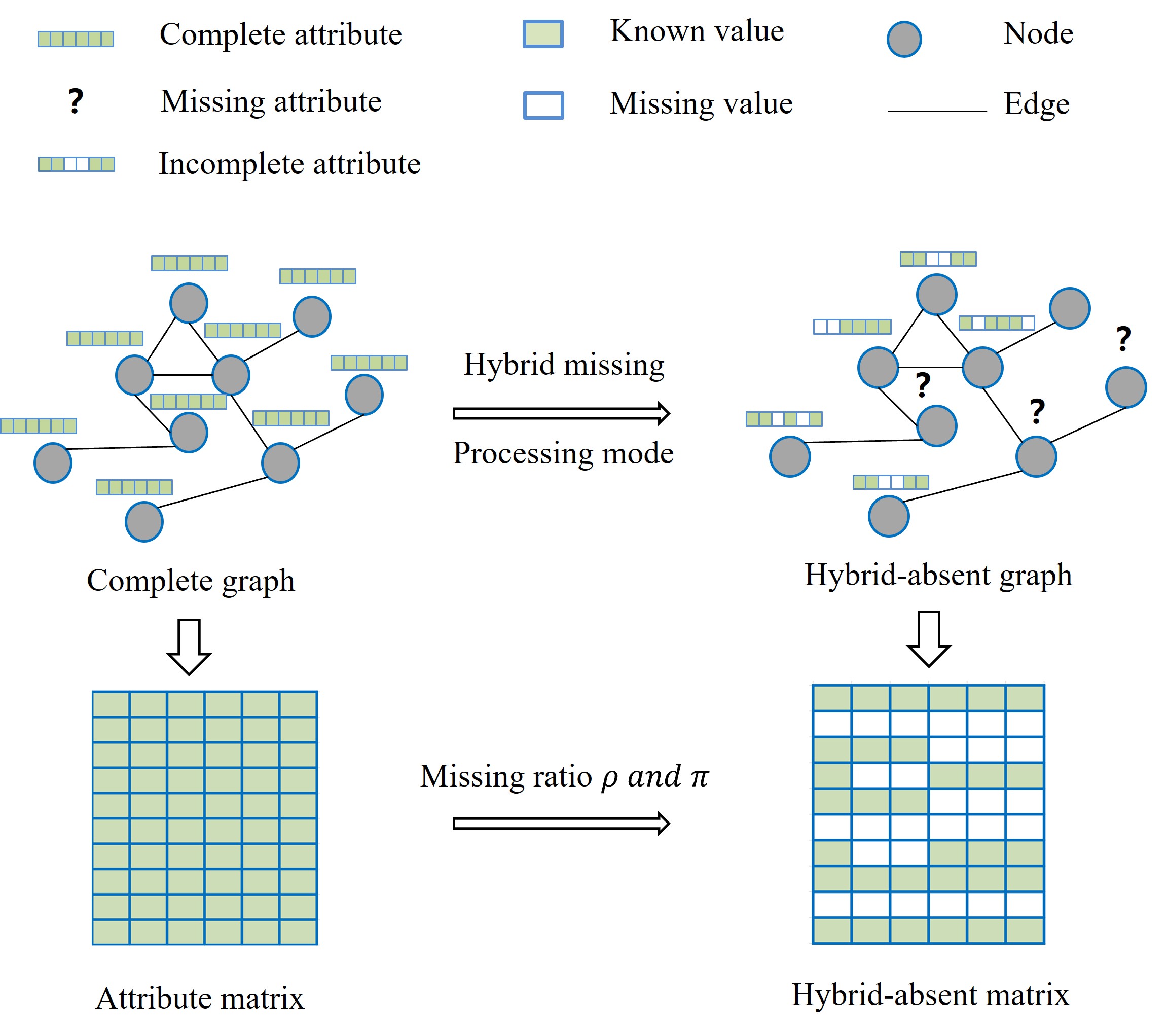}
\caption{The incomplete processing mode of hybrid-absent graphs. }
\label{Fig16}
\end{figure*}

By combining uniform and structural missingness, researchers establish a composite missingness scenario that more accurately reflects the complexities of real-world situations, where data are often incomplete and missing in different ways. A visual representation of this process is provided in Figure \ref{Fig16}.

\subsection{Evaluation index}\label{section6.3.1}
This section presents the tasks addressed in this review and their corresponding evaluation metrics, providing standards to assess algorithm performance. Incomplete graph learning research encompasses multiple tasks, including link prediction, node classification, feature estimation, and node clustering. The following sections provide detailed descriptions.

\subsubsection{Node classification}\label{section6.3.1}
Node classification refers to the task of predicting the node labels in the test set using the node labels, node attributes and the connections between nodes in the training set. For node classification tasks, most methods adopt commonly used evaluation metrics, including Accuracy (ACC), Recall, Precision, $F_1$-score, Macro-$F_1$, and Micro-$F_1$. In this section, we will cover these node classification evaluation metrics related to incomplete graph learning in detail.

ACC refers to the proportion of nodes that are correctly classified, and it is expressed as:
\begin{equation}
\text{ACC} = (\text{TP}+  \text{TN}  )/(\text{TP}+  \text{FP}+  \text{FN}+  \text{TN} ),
\end{equation}
where, TP represents the number of positive samples that are correctly predicted as positive, FP represents the number of negative samples that are incorrectly predicted as positive, TN represents the number of negative samples that are correctly predicted as negative, and FN represents the number of positive samples that are incorrectly predicted as negative.

Recall represents the proportion of correctly identified positive samples to the total number of actual positive samples, and is expressed as:
\begin{equation}
\text{Recall} = \text{TP}  /(\text{TP}+  \text{FN}  ).
\end{equation}

Precision measures the proportion of predicted positive samples that are actually positive, and is expressed as:
\begin{equation}
\text{Precision} = \text{TP}/(\text{TP}+  \text{FP}  ).
\end{equation}

The $F_1$-score is a metric used to measure the accuracy of binary classification models, considering both precision and recall, and is expressed as:
\begin{equation}
F_{1}\text{-}\text{score}    = (\text{2}\times \text{Precision}\times \text{Recall}   )/(\text{Precision } +  \text{Recall} ).
\end{equation}

Macro-$F_1$ is a key metric for evaluating multi-class classification tasks, used to assess the model's overall performance across all categories, and is expressed as:
\begin{equation}
\text{Macro-}F_{1} = \text{2} \times (Precision_{macro}\times  Recall_{macro} )/(Precision_{macro}+   Recall_{macro}),
\end{equation}
where the expressions for $Precision_{macro}$ and $ Recall_{macro}$ are as follows:
\begin{equation}
Precision_{macro} =  {\textstyle \sum_{\text{i}= \text{1}  }^{\text{n} }}Precision_{i}  /n.
\end{equation}
\vspace{-20pt}
\begin{equation}
Recall_{macro} = {\textstyle \sum_{\text{i}= \text{1}  }^{\text{n} }}Recall_{i} /n.
\end{equation}
where $\text{Precision}_{\text{i}} = TP_i/(TP_i +  FP_i)$ and $\text{Recall}_{\text{i}} = TP_i/(TP_i +  FN_i)$.

Micro-$F_1$ score aggregates the contributions of all classes to compute the overall $F_1$ score, particularly useful for class imbalance. In addition, Micro-$F_1$ is a crucial evaluation metric to assess the accuracy and robustness of models in multi-class classification tasks. Its expression is as follows:
\begin{equation}
\text{Micro-}F_{1} = \text{2} \times (Precision_{micro}\times  Recall_{micro}  )/(Precision_{micro}+   Recall_{micro}),
\end{equation}
where the expressions for $Precision_{micro}$ and $ Recall_{micro}$ are as follows:
\begin{equation}
Precision_{micro} = ( {\textstyle \sum_{\text{i}= \text{1}  }^{\text{n} }}TP_{i})/({\textstyle \sum_{\text{i}= \text{1}  }^{\text{n} }}TP_{i}+  {\textstyle \sum_{\text{i}= \text{1}  }^{\text{n} }}FP_{i} ).
\end{equation}
\vspace{-20pt}
\begin{equation}
Recall_{micro} = ({\textstyle \sum_{\text{i}= \text{1}  }^{\text{n} }}TP_{i}   )/({\textstyle \sum_{\text{i}= \text{1}  }^{\text{n} }}TP_{i}+  {\textstyle \sum_{\text{i}= \text{1}  }^{\text{n} }}FN_{i} ).
\end{equation}

Overall, many incomplete graph learning algorithms evaluate performance through node classification tasks, with ACC being the most commonly used evaluation metric. The references corresponding to different evaluation metrics are provided in Table \ref{table8}.

\begin{table}[]
\caption{References corresponding to different evaluation metrics for node classification.}
\label{table8}
\begin{tabular}{p{4cm}|p{11cm}}
\hline
Evaluation metric & References                                                                                  \\ \hline
\multirow{5}{*}{ACC} &\cite{DBLP:journals/nn/SpinelliSU20,DBLP:conf/nips/Morales-Alvarez22,DBLP:conf/aaai/GaoNCTLXZTL23,DBLP:journals/fgcs/TaguchiLM21,10495099,DBLP:conf/aaai/Huo0LHYW23,DBLP:conf/log/RossiK0C0B22,DBLP:conf/icassp/LeiFWQHPY23,Chen_2022,9765782,DBLP:conf/ijcai/TuZLLCZZC22,XIA2024111583,guo2023fairattributecompletiongraph,wang2022schemaguidedeventgraphcompletion,cui2022positionalstructuralnodefeatures,DBLP:journals/fgcs/TaguchiLM21,chen2022wassersteingraphneuralnetworks,um2023confidencebasedfeatureimputationgraphs} \\ \hline
Precision         &\cite{DBLP:conf/nips/Morales-Alvarez22}                                                                                                                                                                \\ \hline
$F_{1} $-score    &\cite{DBLP:conf/nips/Morales-Alvarez22,9514682}                                                                                                                                                        \\ \hline
\multirow{2}{*}{Macro-$F_{1}$}     & \cite{10.1145/3442381.3449914,ZHAO2024122945,9724614,LI2023424,DBLP:journals/kbs/ChenL24,wang2022deeppartialmultiplexnetwork,DBLP:conf/aaai/LuSH019}                                                                                                                                          \\ \hline
\multirow{2}{*}{Micro-$F_{1} $}    & \cite{ZHAO2024122945,9724614,LI2023424,DBLP:journals/kbs/ChenL24,wang2022deeppartialmultiplexnetwork,DBLP:conf/aaai/LuSH019}                                                                                                                 \\ \hline
Recall     &\cite{DBLP:conf/nips/Morales-Alvarez22}                                                                   \\ \hline
\end{tabular}

\end{table}

\subsubsection{Node clustering}\label{section6.3.2}
Node clustering involves grouping similar nodes based on the similarities and connection patterns between them. For evaluating node clustering, metrics such as Normalized Mutual Information (NMI), Clustering Accuracy (AC), Purity, and Adjusted Rand Index (ARI) are used. In this section, we will cover these node clustering evaluation metrics related to incomplete graph learning in detail.

NMI is a metric used to evaluate the effectiveness of clustering or the similarity between two clustering results. It is based on the concept of Mutual Information (MI) and quantifies the correlation or similarity between two sets by calculating the difference between their joint distribution and their independent distributions. Its expression is as follows:
\begin{equation}
\text{NMI}(Y, C) = 2 \times I(Y; C)/[H(Y) + H(C)],
\end{equation}
where $Y$ denotes the true categories, $C$ denotes the clustering results, $\text{H}\left ( \cdot  \right ) $ represents entropy, and $I(Y;C)$ represents mutual information.

AC is an important metric for evaluating clustering performance. It is calculated based on the degree of match between the clustering results and the true labels, and is expressed as:
\begin{equation}
\text{AC} =\sum_{i=1}^{n} {\delta (s_i, \text{map}(r_i))}/{\text{n} },
\end{equation}
where $r_{i} $ represents the labels assigned after clustering, $s_{i} $ represents the true labels, $n$ represents the number of data instances, and 
$\delta $ denotes the indicator function, characterized by the following formula:
\begin{equation}
\delta (x,y) =
\begin{cases}
1 & \text{if } x = y \\
0 & \text{otherwise}.
\end{cases}
\end{equation}

Purity is a clustering evaluation metric used to estimate the proportion of nodes from the dominant class within each cluster, as well as the overall proportion of these dominant-class nodes across all clusters. It reflects the consistency between the clustering result and the true classes, and is specifically defined as follows:
\begin{equation}
Purity = \sum_{i=1}^{j}(n_i \times p_i)/{n},
\end{equation}
where $j$ and $n$ are the number of clusters and nodes, respectively. $n_i$ is the number of nodes in the $i$-th cluster, $p_i$ indicates the proportion of the dominant class nodes within the $i$-th cluster.

ARI quantifies the concordance between two clustering outcomes by enumerating the pairs of samples that are either grouped together or separated across distinct clusters. The formula for its computation is presented below:
\begin{equation}
\text{ARI}= \left [ RI- E\left ( RI \right )  \right ]/\left [ max\left ( RI \right ) - E\left ( RI \right )  \right ]  ,
\end{equation}
where $RI$ is rand index, $E(RI)$ is the expected value of the $RI$ under random clustering, $max(RI)$ is the maximum possible value of the $RI$. The calculation formula for $RI$ is as follows:
\begin{equation}
\text{RI}=\left ( a+  d \right ) /\left ( a+  b+  c+  d \right )   ,
\end{equation}
where $a$ denotes the count of point pairs that are co-clustered in both the ground truth and the experimental scenarios, $b$ denotes those that belong to the same cluster in the true condition but not in the experimental condition, $c$ represents pairs that do not belong to the same cluster in the true condition but do in the experimental condition, and $d$ represents pairs that do not belong to the same cluster in either condition. The ARI ranges from -1 to 1, with higher values indicating a closer match to the true clustering, reflecting better clustering performance.

Overall, many incomplete graph learning algorithms evaluate performance through node clustering tasks. NMI and AC are the most commonly used evaluation metric. The corresponding references for different evaluation metrics are provided in Table \ref{table9}.

\begin{table}[]
\caption{References corresponding to different evaluation metrics for node clustering.}
\label{table9}
\begin{tabular}{p{4cm}|p{11cm}}
\hline
Evaluation metric & References                                                                                  \\ \hline
\multirow{5}{*}{NMI}              &  \cite{9765782,DBLP:journals/kbs/WangYLF19,DBLP:journals/tsmc/WenZFZXZL23,DBLP:conf/ijcai/WangZLYZ19, DBLP:conf/aaai/GuoY19,DBLP:journals/tkde/LiangYX23, DBLP:journals/www/HeZCW23,DBLP:journals/tmm/WenYZXWFZ21,ZHANG2022108412,9724614,LI2023424,DBLP:journals/kbs/ChenL24,wang2022activeaugmentationfreegraphcontrastivelearning,2020One2Multi,DBLP:conf/aaai/LuSH019} \\ \hline
\multirow{3}{*}{AC}         & \cite{9765782,DBLP:journals/kbs/WangYLF19,DBLP:journals/tsmc/WenZFZXZL23,DBLP:conf/ijcai/WangZLYZ19, DBLP:conf/aaai/GuoY19,DBLP:journals/tkde/LiangYX23, DBLP:journals/www/HeZCW23,DBLP:journals/tmm/WenYZXWFZ21,ZHANG2022108412,wang2022deeppartialmultiplexnetwork,wang2022activeaugmentationfreegraphcontrastivelearning}                                                                                                                                                                \\ \hline
\multirow{2}{*}{Purity }   & \cite{DBLP:journals/tsmc/WenZFZXZL23,DBLP:journals/tkde/LiangYX23, DBLP:journals/www/HeZCW23,DBLP:journals/tmm/WenYZXWFZ21,ZHANG2022108412}                                                                                                                                                        \\ \hline
ARI     & \cite{9724614,LI2023424,DBLP:journals/kbs/ChenL24,2020One2Multi}                                                                                                                                          \\ \hline

\end{tabular}

\end{table}

\subsubsection{Link prediction}\label{section6.3.3}
Link prediction can be viewed as a classification task that aims to predict the existence of unknown links based on the known links in a graph. Commonly used evaluation metrics for link prediction in incomplete graphs include Area Under the Curve (AUC) \cite{DBLP:journals/fgcs/TaguchiLM21,Chen_2022,9765782,10210121,zhang2022completingnetworkslearninglocal,DBLP:journals/fgcs/TaguchiLM21,um2023confidencebasedfeatureimputationgraphs,DBLP:conf/aaai/LuSH019} and Average Precision (AP) \cite{9765782,10210121,um2023confidencebasedfeatureimputationgraphs}.

In the link prediction task, AUC evaluates the model's ability to rank a randomly selected missing edge higher than a non-existent edge. Specifically, randomly select a missing edge and a fictional edge, and compare the scoring situations of these two types of edges. In $m$ comparisons, if there are $m^{'}$ times when the score of the fictional edge is lower than that of the missing edge, and there are $m^{''}$ times when the scores of the two are the same. The expression for AUC is:
\begin{equation}
\text{AUC} =(m^{'} +  \text{0.5} m^{''}  )/{\text{m} },
\end{equation}
when the value of AUC is closer to 1, the accuracy of link prediction is higher.

The expression for AP is:
\begin{equation}
\text{AP} =  \text{P} \left ( \text{Re}  \right )/\text{M}, 
\end{equation}
where $\text{P}(\text{Re})$ represents the precision value for the relevant set $\text{Re}$. A higher value of AP, closer to 1, indicates better accuracy in link prediction.

\subsubsection{Feature estimation}\label{section6.3.4}
Feature estimation indicates the process of predicting or estimating missing attributes in a graph. For feature estimation tasks, Mean Absolute Error (MAE), Root Mean Squared Error (RMSE), Mean Absolute Percentage Error (MAPE), Recall, and Normalized Discounted Cumulative Gain (NDGG) are commonly used to evaluate the performance of incomplete graph learning methods. The corresponding references for different evaluation metrics are provided in Table \ref{table10}.

\begin{table}[]
\caption{References corresponding to different evaluation metrics for feature estimation.}
\label{table10}
\begin{tabular}{p{4cm}|p{11cm}}
\hline
Evaluation metric & References                                                                                  \\ \hline
MAE              &  \cite{DBLP:conf/aaai/ZhongGY23} \\ \hline
\multirow{2}{*}{RMAE}        & \cite{2017Graph,zhang2020inductivematrixcompletionbased,DBLP:conf/nips/MontiBB17,2018Convolutional,DBLP:conf/cikm/0005ZTZHD021,
DBLP:journals/iotj/WuXFW22,yoo2023accuratenodefeatureestimation}                                                                                                                                                                \\ \hline
MAPE    & \cite{DBLP:journals/kbs/KongZSZLY23,DBLP:journals/corr/abs-2406-03511}                                                                                                                                                        \\ \hline
Recall     & \cite{yoo2023accuratenodefeatureestimation,zhang2023amgclfeatureimputationattribute,10210121,tu2023revisitinginitializingrefiningincomplete}                                                                                                                                          \\ \hline
\multirow{2}{*}{NDCG}     & \cite{yoo2023accuratenodefeatureestimation,zhang2023amgclfeatureimputationattribute,10210121,tu2023revisitinginitializingrefiningincomplete}, \cite{yoo2023accuratenodefeatureestimation}                                                                                                                                         \\ \hline
\end{tabular}

\end{table}

MAE is a regression error metric that calculates the average of the absolute differences between predicted and ground truth values. Its formula is as follows:
\begin{equation}
\text{MAE} = \sum_{k=1}^{n} |y_k - \hat{y}_k|/n, 
\end{equation}
where $n$ denotes the quantity of nodes, $y_{k}$ denotes the ground truth value, and $\hat{y}_k$ denote the predicted value for the k-th node.

RMSE is the square root of the mean squared error, commonly used to assess the accuracy of prediction models. The corresponding equation is presented below:
\begin{equation}
\text{RMSE} = \sqrt{ \sum_{j=1}^{n} (y_j - \hat{y}_j)^2/n},
\end{equation}
where $n$ denotes the quantity of observations, $y_{j}$ denotes the ground truth of the $j$-th observation, and $\hat{y}_j$ denotes the predicted value of the $j$-th observation. A smaller RMSE indicates higher prediction accuracy.

MAPE quantifies the mean of absolute deviations between predicted and actual values, represented as a percentage relative to the ground truth values. It reflects the average prediction error relative to the true values. The corresponding equation is presented below:
\begin{equation}
\text{MAPE}(X, \hat{X}) =\left [ \left( \sum_{i=1}^{m} \left| (X_i - \hat{X}_i)/{X_i} \right| \right)/m \right ]  \times 100\%,
\end{equation}
where $m$ denotes the quantity of nodes.

For the feature estimation task, the expression for recall is given as follows:
\begin{equation}
\text{Recall} = \left | \text{R}\cap \text{T} \right |   /{\left | \text{T} \right |  },
\end{equation}
where $\text{R}$ denotes the set of attributes predicted by the model as relevant, and $\text{T}$ represents the set of true relevant attributes in the test set. The overall recall for the dataset is obtained by calculating the recall for each node and averaging the values across all nodes.

NDCG serves as a metric for evaluating the ranking quality of node attributes in the recommendation list, and its calculation formula is as follows:
\begin{equation}
\text{NDGG}=\text{DGG} /\text{IDGG} ,
\end{equation}
where DCG stands for discounted cumulative gain, and the specific formula is:
\begin{equation}
\text{DGG}=  \sum_{\text{j}= \text{1}  }^{\text{k} } \left [ rel_{j} /{\log_{2}{\left ( j+  \text{1} \right )   } } \right ],
\end{equation}
where k represents the size of the attribute list to be recommended, and $rel_{j}$ denotes the relevance score of the attribute recommended at position $j$ in the node attribute vector.
IDCG stands for ideal discounted cumulative gain, which represents the ideal case of discounted cumulative gain. The specific formula is:
\begin{equation}
\text{IDGG}=\sum_{j= 1}^{\left | REL \right | }\left [ rel_{j} / log_{2}\left ( j+1 \right )   \right ],
\end{equation}
where REL refers to the originally recalled set of attributes, which is sorted by relevance score. The NDCG score spans from 0 to 1, with values nearer to 1 signifying a higher level of ranking quality.

\subsection{Applications}\label{section6.4}
Incomplete graphs are prevalent across various fields, and with the continuous efforts of researchers, an increasing number of promising methods for incomplete graph learning have emerged, spanning multiple application domains. Sections \ref{section3}, \ref{section4}, and \ref{section5} primarily focus on relevant learning methods based on different types of incomplete graphs. To encourage further attention from researchers on the issues related to incomplete graphs in diverse fields and to expand the application scope of incomplete graph learning, this section will focus on the current hot application areas in incomplete graph research, specifically including knowledge graphs, transportation systems, and recommendation systems, which offer substantial practical value. The details of these applications are as follows.

\subsubsection{Knowledge graphs}\label{section6.4.1}
A Knowledge Graph (KG) is a structured knowledge base that represents entities, concepts, and their relationships using a graph format. In recent years, researchers have discovered that knowledge graphs also suffer from incomplete attributes, where nodes possess numerical attributes, but the values of these attributes are often missing. In response to this issue, Kotnis and Garcia-Duran~\cite{kotnis2019learning} proposed a two-step framework called NAP++. The framework first extends the knowledge graph representation method to learn node representations for a KG enriched with numerical node attributes. It then constructs a k-nearest
neighbor graph based on these representations to propagate the observed node attributes to those that are missing. In response to the limitations of propagating information through surrogate graphs constructed from embeddings, Bayram et al.~\cite{DBLP:conf/icassp/BayramG021} proposed the Multi-Relational Attribute Propagation (MRAP) model, which directly leverages the underlying structure of the KG. Xue et al.~\cite{DBLP:conf/semweb/XueLZ22} introduced several novel methods for exploring and utilizing the rich semantic knowledge of language models in attribute prediction. Furthermore, to fully utilize multimodal data for completing missing attributes in electronic products, Wang et al.~\cite{DBLP:conf/www/WangSZCH23} proposed a robust three-stream framework called MPKGAC. This model first constructs a multimodal product KG using multimodal features and then transforms the attribute completion problem into a multimodal KG completion task.

In summary, although some researchers have recognized the problem of missing attributes in KGs, current exploration of this problem and utilization of observable attributes remain insufficient. Given the importance of attribute information, we anticipate that more researchers will focus on this area in the future.

\subsubsection{Transportation systems}\label{section6.4.2}
With the rapid advancement of Intelligent Transportation Systems (ITS), vast amounts of traffic data such as vehicle trajectories, traffic flow, signal states, road occupancy, and weather conditions are continuously generated and analyzed. These datasets serve as the foundation for critical applications like traffic planning, congestion management, accident prevention, and public transport optimization. However, the absence of traffic data, often caused by factors such as equipment failures, network disruptions, data entry mistakes, and privacy regulations, remains a common challenge. This significantly hinders the effective utilization and in-depth analysis of the available data.

Given the significant impact of road network structures on traffic conditions, recent research has increasingly approached the problem of missing traffic data as an incomplete graph modeling problem. These studies employ GNN techniques and have produced promising results in the field of traffic prediction~\cite{DBLP:conf/iclr/LiYS018, DBLP:conf/ijcai/YuYZ18, DBLP:conf/ijcai/WuPLJZ19, DBLP:conf/aaai/ZhengFW020}. For example, Zhang et al.~\cite{ZHANG2021103372} presented a deep learning model designed to combine online traffic data imputation and prediction at the network level. This model integrates both data imputation and recurrent neural networks to estimate the anticipated results for data imputation and traffic prediction tasks. Yao et al.~\cite{DBLP:journals/tits/YaoGZMWL21} examined the graph structure of spatial flows and introduced a spatial interaction-based GCN model to estimate spatial origin-destination flows. To address the limitation that a static, distance-based graph fails to capture temporal variations in spatial correlations, Xu et al.~\cite{DBLP:journals/tits/XuPWSL22} proposed an imputation learning model to impute missing data, which uses a GNN to aggregate spatiotemporal information from a graph built from correlation coefficients derived from historical data. Zhong et al.~\cite{DBLP:conf/icdcs/ZhongSJZS21} proposed a GCN model based on heterogeneous graphs, which constructs a multigraph from geographical and historical data to explicitly model the dependencies between road segments. The model imputes missing values through a recurrent process, which is seamlessly embedded within the prediction framework. Kong et al.~\cite{DBLP:journals/kbs/KongZSZLY23} introduced a novel graph generation model that leverages recurrent input data and historical information to model dynamic spatial correlations between road network nodes at each time step. They employ a dynamic graph convolutional gated recurrent unit to apply graph convolution to both static and dynamic graphs, thereby capturing temporal and spatial dependencies in the data more effectively. Moreover, Xu et al.~\cite{DBLP:conf/cvpr/XuBCCF23} proposed a unified Graph-based Conditional Variational Recurrent Neural Network (GC-VRNN) for trajectory and traffic prediction. This method introduces a multi-space GNN to extract spatial attributes from incomplete trajectory data and employs a conditional variational recurrent neural network to capture temporal dependencies and missing patterns.
Marisca et al.~\cite{DBLP:conf/nips/MariscaCA22} proposed an attention-based architecture that learns spatiotemporal representations from sparse discrete observations. The model exploits a spatiotemporal propagation mechanism aligned with the imputation task to reconstruct missing observations for a given sensor and its neighboring nodes. 

Although GNNs have shown great promise in predicting incomplete traffic data, there is still considerable room for improvement. For example, how can multi-source data from various sources, such as traffic sensors, weather stations, and social media platforms, be leveraged to impute missing values? Future research could focus on effectively integrating these multi-source data sources and dynamically modeling the interactions between different spatiotemporal components (e.g., roads, intersections, or vehicle types) to improve prediction accuracy. In addition, current models often lack interpretability. Future efforts could aim to develop GNN models for incomplete traffic data that not only provide accurate predictions, but also offer insights into the underlying causes of these predictions. This would improve the reliability of data imputation and support decision-making in traffic management.

\subsubsection{Recommendation systems}\label{section6.4.3}
Recommendation systems provide personalized content or product recommendations by analyzing user behaviors and preferences. However, in practical applications, user-item interaction data is often incomplete, with a significant number of missing attributes. This incompleteness undermines the accuracy of predictions and the overall performance of recommendation systems. To address this issue, matrix completion becomes a critical method for recovering missing data in recommendation systems. For instance, the GC-MC~\cite{2017Graph}, IGMC~\cite{zhang2020inductivematrixcompletionbased}, GRAPE~\cite{2020Handling}, RGCNN~\cite{DBLP:conf/nips/MontiBB17}, CGMC~\cite{2018Convolutional}, IMC-GAE~\cite{DBLP:conf/cikm/0005ZTZHD021}, and ROGMC~\cite{DBLP:journals/corr/abs-2403-04504} address the problem of incomplete matrix completion in recommendation systems. Most of these methods construct a bipartite graph from the user-item rating matrix, where users are nodes on one side, items are nodes on the other side, and either ratings or interactions serve as the edges connecting the two sides. Consequently, the matrix completion problem can be framed as an edge prediction problem on an incomplete bipartite graph.

Most current methods for incomplete graph learning in recommendation systems focus on matrix completion, which aims to fill in missing entries in the user-item interaction matrix to improve recommendation quality. However, this approach has limitations, as it overlooks the significant issue of incomplete attribute information for users or items. User or item attribute information, such as age, gender, preferences, categories, descriptions, and evaluations, is crucial for building accurate and efficient recommendation systems. However, in practice, due to privacy policies, technical challenges, and data collection costs, attribute information is frequently absent or incomplete. Given the widespread issue of missing attribute information, future research should focus on effectively handling this incompleteness and explore ways to optimize recommendation system performance with incomplete data. 

\subsubsection{Summary and discussion}\label{section6.4.4}
Beyond these domains, the issue of incomplete graphs affects many other fields. For example, Liu et al.~\cite{10195200} proposed the Missing Event-Aware Temporal GNN (MTGN) for event prediction with some events missing, which simultaneously models the evolving graph structure and event timings, enabling dual predictions of future events and their occurrence times. Cheng et al.~\cite{DBLP:conf/aaai/ChengZTG024} addressed the problem of detecting rumor sources in the presence of incomplete user data and proposed a novel approach. These significant developments not only provide novel insights for ongoing research endeavors but also demonstrate the potential for diverse domains to develop effective solutions to the challenge of incomplete attributes, thereby driving further development and application of related technologies. We expect that researchers will continue to tackle related challenges and propose more methods for incomplete graph learning, expanding their applications into broader and more impactful domains.
\section{Discussion and conclusion}\label{section7}

The aforementioned content offers a comprehensive survey of the present state of research and methodologies in incomplete graph learning. In this section, we explore several potential future research directions. Additionally, we offer a comprehensive summary, aiming to provide a solid foundation and practical guidance for future researchers.

\subsection{Discussion}\label{section7.1}
Incomplete graph learning initially gained attention due to the widespread presence of missing attributes in real-world graphs. Since the emergence of incomplete graph learning, remarkable advancements have been achieved in the methods and applications of incomplete graphs, as discussed in Sections \ref{section3} to \ref{section6}. However, several issues and challenges remain in this research area. The following discusses potential research directions to address these challenges.

\subsubsection{Interpretability and robustness}\label{section7.1.1}

Incomplete graphs may be risk-sensitive and privacy-related (e.g., in social networks where certain users are reluctant to offer personal information), rendering an interpretable and robust graph learning approach crucial for adapting to such learning scenarios. However, most existing incomplete graph learning methods focus on achieving higher performance on downstream tasks through black-box deep learning models, neglecting the interpretability~\cite{DBLP:journals/nn/YangWHZ24, DBLP:journals/nn/ZhangWHQHG25} of learned representations and prediction outcomes. In addition, while most methods acknowledge the incompleteness of input data, existing approaches lack robustness, as they are typically designed to handle only a single type of incomplete graphs, making them unsuitable for more complex scenarios involving multiple attribute deficiencies. Consequently, exploring interpretable and robust incomplete graph learning methods represents an intriguing and practical direction for future research, which can enhance the reliability and effectiveness of graph learning methods in real-world applications.

\subsubsection{Learning More Complex Graphs}\label{section7.1.2}
Most current research in incomplete graph learning focuses on simpler graph structures, such as homogeneous or homophilic graphs, while research on more complex graph types, including heterogeneous~\cite{10.1145/3442381.3449914}, dynamic~\cite{DBLP:journals/tkdd/EkleE24}, and spatiotemporal graphs~\cite{DBLP:journals/kbs/KongZSZLY23}, as well as class-imbalanced graphs~\cite{DBLP:journals/corr/abs-2403-04468}, remains underexplored. Given the widespread prevalence of these complex graph types and their vulnerability to attribute incompleteness, addressing attribute missingness in these graphs is both crucial and timely.

For complex graphs, the key challenge lies in developing efficient incomplete graph learning methods that can effectively capture and represent their unique characteristics. While several approaches attempt to leverage topological structure for incomplete graph learning~\cite{10.1145/3442381.3449914}, their effectiveness is limited when dealing with complex graphs. Therefore, developing methods for incomplete complex graphs is an urgent research direction that needs to be addressed.

\subsubsection{ Learning with multiple pretext tasks}\label{section7.1.3}
Most of the current research in incomplete graph learning focuses on node-level tasks such as node classification, node clustering, and node attribute completion, and these efforts have laid a solid foundation for graph analysis. However, the applications of graph data span a wide range, going beyond node-level tasks to encompass other important graph tasks, such as edge classification and graph classification~\cite{DBLP:journals/nn/LiLGLQW25, DBLP:journals/nn/YuMBZWDH23}. Unfortunately, effective solutions for these tasks are still lacking in the domain of incomplete graph learning.

Given the diversity of downstream tasks in real-world graph data, expanding the scope of incomplete graph learning tasks and enhancing their generalization capabilities are of paramount importance. This not only facilitates a more comprehensive understanding and utilization of graph data but also markedly enhances the performance of incomplete graph learning methods in practical scenarios.

\subsubsection{Broader scope applications}\label{section7.1.4}
Graphs, as a fundamental data structure, are widely prevalent across various domains. However, in most current application areas, the critical issue of incomplete graphs is seldom fully addressed, significantly limiting the broader adoption of incomplete graph learning techniques. Currently, practical applications are concentrated in a few select domains, such as recommender system optimization, knowledge graph completion, and social network analysis, with limited exploration of other potential application areas. Only a small group of pioneering researchers have begun to explore emerging fields, such as traffic flow prediction. In light of this, broadening the application of incomplete graph learning to domains such as financial network analysis~\cite{DBLP:conf/www/ZhangLZ24}, federated recommendation~\cite{DBLP:conf/kdd/ZhangL0ZYY24} and medical networks~\cite{DBLP:journals/corr/abs-1904-00326} represents a highly promising endeavor.

\subsubsection{Combining with large language models}\label{section7.1.5}
With the successful application of Large Language Models (LLMs)~\cite{DBLP:conf/ijcai/ChavanMKDG24} in fields such as natural language processing, graph learning~\cite{DBLP:journals/corr/abs-2408-08685}, and computer vision, integrating them with incomplete graph learning has become a promising research direction with significant potential. LLMs, with their powerful representational capacity, can learn complex features and contextual information, helping to complete missing attributes in incomplete graphs and improving the accuracy and robustness of graph learning. Future research could explore how to effectively integrate LLMs with incomplete graph learning to better address challenges in this domain. Currently, there is a lack of studies combining LLMs with incomplete graph learning, making this integration not only a new approach for handling incomplete graphs but also opening up broad prospects for the further development of incomplete graph learning.

\subsection{Conclusion}\label{section7.2}
Incomplete graph learning is still an emerging and promising field of research. A key challenge is designing effective methods for the practical application of incomplete graph learning.

We have gathered information from various publications and organized it into a cohesive framework, aiming to provide a comprehensive review of incomplete graph learning. In this review, we presented the fundamental concepts of incomplete graphs and classified the incomplete graph learning methods into three categories based on the type of incomplete graphs. Based on the types of incomplete graph learning, we present a detailed overview of the relevant work in this field to elucidate the characteristics, application scenarios, and advantages and disadvantages of each approach. Additionally, we provide an overview of the datasets, incomplete processing modes, evaluation tasks, and applications. Finally, we highlight the key open problems and propose directions for future exploration. To the best of our knowledge, this is the first review on incomplete graph learning, which is one of the most important research fields in graph learning since many graphs are missing their attributes in the real world. We hope this review provides a thorough and comprehensive summary of the latest advancements, challenges, and future research directions in incomplete graph learning, benefiting both academic and industrial audiences.

\printcredits
\section*{Declaration of competing interest}
The authors declare that they have no known competing financial interests or personal relationships that could have appeared to influence the work reported in this paper.
\section*{Acknowledgments}
This work was supported by the National Key R\&D Program of China [grant number 2021ZD0112500]; the National Natural Science Foundation of China [grant numbers 62202200, 62402197]; the fund of Supporting the Reform and Development of Local Universities (Disciplinary Construction) and the special research project of First-class Discipline of Inner Mongolia A. R. of China under Grant YLXKZX-ND-036. We would like to thank the reviewers for their valuable comments.

\bibliographystyle{newapa}

\bibliography{cas-refs}



\end{document}